\pdfoutput=1

\documentclass[11pt]{article}

\usepackage[final]{acl}

\usepackage{times}
\usepackage{latexsym}

\usepackage[T1]{fontenc}

\usepackage[utf8]{inputenc}

\usepackage{microtype}

\usepackage{inconsolata}

\usepackage{graphicx}

\usepackage{hyperref}       
\usepackage{url}            
\usepackage{booktabs}       
\usepackage{amsfonts}       
\usepackage{nicefrac}       
\usepackage{xcolor}         
\usepackage[flushleft]{threeparttable}
\usepackage{wrapfig}
\usepackage{tcolorbox}
\usepackage{diagbox}
\usepackage{arydshln}
\usepackage{multirow}
\usepackage{anyfontsize}
\usepackage{amsmath}
\usepackage{makecell}
\usepackage{float}
\usepackage{pifont}
\newcommand{\cmark}{\ding{51}}%
\newcommand{\xmark}{\ding{55}}%
\usepackage{enumitem}
\usepackage{spverbatim}
\usepackage{fancyvrb}
\usepackage{fvextra}
\usepackage{xcolor}

\definecolor{light-gray}{gray}{0.95}
\newcommand{\code}[1]{\colorbox{light-gray}{\texttt{#1}}}

\title{Synth-SBDH: A Synthetic Dataset of Social and Behavioral Determinants of Health for Clinical Text}

\author{
 \textbf{Avijit Mitra\textsuperscript{1,2}},
 \textbf{Zhichao Yang\textsuperscript{1,2}},
 \textbf{Emily Druhl\textsuperscript{2}},
 \textbf{Raelene Goodwin\textsuperscript{2}},
 \textbf{Hong Yu\textsuperscript{1,2,3,4}}
\\
\\
 \textsuperscript{1}Manning College of Information and Computer Sciences, University of Massachusetts Amherst,\\
 \textsuperscript{2}U.S. Department of Veterans Affairs, \\
 \textsuperscript{3}Department of Medicine, University of Massachusetts Chan Medical School,\\
 \textsuperscript{4}Miner School of Computer and Information Sciences, University of Massachusetts Lowell
\\
 \texttt{avijitmitra@umass.edu} \hspace{2mm} 
}

\begin{document}
\maketitle
\begin{abstract}
Social and behavioral determinants of health (SBDH) play a crucial role in health outcomes and are frequently documented in clinical text. Automatically extracting SBDH information from clinical text relies on publicly available good-quality datasets. However, existing SBDH datasets exhibit substantial limitations in their availability and coverage. In this study, we introduce Synth-SBDH
\footnote{\href{https://github.com/avipartho/Synth-SBDH}{https://github.com/avipartho/Synth-SBDH}}
, a novel synthetic dataset with detailed SBDH annotations, encompassing status, temporal information, and rationale across 15 SBDH categories. We showcase the utility of Synth-SBDH on three tasks using real-world clinical datasets from two distinct hospital settings, highlighting its versatility, generalizability, and distillation capabilities. Models trained on Synth-SBDH consistently outperform counterparts with no Synth-SBDH training, achieving up to 63.75\% macro-F improvements. Additionally, Synth-SBDH proves effective for rare SBDH categories and under-resource constraints while being substantially cheaper than expert-annotated real-world data. Human evaluation reveals a 71.06\% Human-LLM alignment and uncovers areas for future refinements. 
\end{abstract}

\section{Introduction}

\label{intro}
In healthcare research, understanding the nuanced interplay between social and behavioral determinants of health (SBDH) and health outcomes is imperative for improving patient care and population health management. `SBDH' encompasses both social determinants of health (SDOH) such as housing instability, job insecurity, food insecurity, etc. and behavioral factors such as drug abuse, tobacco use, etc., which are widely recognized for their significant impact on individual's physical and mental health \citep{franke2014toxic,nelson2020adversity,turner2000social,shonkoff2012committee}. There is a growing body of research to link SBDH with various adverse outcomes such as opioid overdose \citep{dasgupta2018opioid,volkow2021changing}, suicide mortality \citep{blosnich2020social,mitra2023associations,kposowa2001unemployment}, Alzheimer's disease \citep{majoka2021effect,adkins2023structural}, etc. Moreover, SBDH extraction from clinical notes has been shown to be instrumental in predicting hospital admissions, suicide attempts, and suicide deaths, among others \citep{chen2020social,nijhawan2019clinical,mitra2025post,takahashi2015health}.

\begin{table*}
  \fontsize{8pt}{10pt}\selectfont
  \centering
  \begin{threeparttable}
  \begin{tabular}{lllllll}
    \toprule
    Dataset     & \#Samples     & Source     & Labeller    & \#Categories & Behavioral & Rationale \\
    \midrule
    MIMIC-SBDH \citep{ahsan2021mimic}  & 7,025        & MIMIC-III   & Human   & 7     & \cmark & \xmark \\
    SHAC \citep{lybarger20232022}\tnote{$\dagger$}      & 4,405        & MIMIC-III+UW\tnote{*}  & Human & 5  & \cmark    & \xmark \\
    SDOH-NLI \citep{lelkes2023sdoh}    & 1,398        & MTSamples  & Human    & 10    & \cmark & \xmark       \\
    PedSHAC \citep{fu2024extracting}\tnote{$\dagger$}    & 1,260        & UW\tnote{*} & Human  & 10  & \cmark    & \xmark  \\
    \citet{guevara2024large}   & 5,328        & MIMIC-III   & Human   & 6    & \xmark & \xmark       \\
    \citet{guevara2024large}   & 480        & GPT-3.5   & Human   & 6    & \xmark & \xmark  \\
    \midrule
    Synth-SBDH (Ours)  & 8,767        & GPT-4      & GPT-4+Human    & 15    & \cmark & \cmark\\
    \bottomrule
  \end{tabular}
  \begin{tablenotes}
    \item[$\dagger$] Not released at the time of this writing.
    \item[*] University of Washington.
  \end{tablenotes}
  \end{threeparttable}
  \caption{Comparison of publicly available SBDH datasets. Synth-SBDH includes both social and behavioral determinants, covering 15 categories. In addition, Synth-SBDH provides annotation rationales.}
  \label{dataset_comp}
  \vspace{-2mm}
\end{table*}

While SBDHs are primarily documented in electronic health records (EHRs) through structured data and unstructured clinical notes, the former has been the dominant source for relevant research. However, structured data sources lack comprehensiveness and reliability \citep{truong2020utilization,heidari2023z}, posing significant challenges to research and clinical care. Instead, SBDHs are most frequently described in the free text of EHR notes with nuanced descriptions that are unavailable through structured sources. Research indicates that unstructured notes may contain up to 90 times more SBDH information than their structured counterparts \citep{dorr2019identifying}. Natural language processing (NLP) can address these challenges by automating their extraction from clinical notes \citep{ahsan2021mimic, lybarger2023leveraging}. However, such automation requires robust datasets for model training. Existing SBDH datasets suffer from notable limitations such as their restricted public availability, lack of relevant information, etc., hindering progress in this domain. Consequently, the development of comprehensive and high-quality SBDH datasets is paramount for advancing NLP in healthcare.

One major challenge for releasing data based on EHRs lies in the sensitive nature of the data, which limits their availability to the broader research community. Additionally, human chart review of EHR notes on a large scale is not only time-consuming but also prohibitively expensive. As a result, most publicly available SBDH datasets derive from the same data sources and contain limited examples (Table \ref{dataset_comp}). In contrast, with the rapid advancements in large language models (LLM), LLMs have shown remarkable performance across different domains including healthcare research \citep{singhal2023large,singhal2023towards,thirunavukarasu2023large,nori2023capabilities}. In particular, LLMs have been utilized for generating patient-physician dialogue from notes \citep{wang2024notechat} and synthetic clinical data \citep{li2023two,guevara2024large}. 

Addressing the limitations of current SBDH datasets and acknowledging the potential of LLMs in healthcare, this study introduces Synth-SBDH, a novel synthetic SBDH dataset that mimics EHR notes. To the best of our knowledge, Synth-SBDH is the largest publicly available SBDH dataset (Table \ref{dataset_comp}), comprising 8,767 examples generated and annotated by an LLM (GPT-4 \citep{achiam2023gpt}) with detailed SBDH information, encompassing various dimensions such as presence, temporality, and rationale across 15 meticulously chosen SBDH categories. While Synth-SBDH is smaller than many contemporary synthetic datasets \citep{wang2022self,wang2024notechat,kweon2023publicly}, we argue that a high-quality synthetic dataset can lead to a substantial gain in performance, particularly due to the imbalanced nature of real-world clinical data \citep{guevara2024large,li2023two}. Moreover, expert review of a smaller dataset is faster and mitigates privacy concerns associated with real-world EHR data while also limiting computational costs and the expense of manual review. Therefore, we conduct an expert evaluation of the Synth-SBDH test split, providing a valuable resource for training and fair evaluation of NLP models in the SBDH extraction task.

We evaluate the utility of Synth-SBDH across multiple NLP tasks leveraging real-world datasets. Specifically, we highlight three key aspects of Synth-SBDH: versatility, generalizability, and distillation capabilities. Our extensive evaluations demonstrate its potential to enhance a broad range of models' performance on SBDH detection. In summary, our contributions can be summarized as follows:
\begin{enumerate}
    \item We introduce Synth-SBDH, the largest publicly available SBDH dataset, encompassing 15 SBDH categories with detailed annotations. A comprehensive human evaluation demonstrates 71.06\% human-LLM alignment for SBDH annotations and a rating of 3.66 on a scale of 4 for annotation rationales. 
    \item We show that models with different architectural backbones when trained on Synth-SBDH, exhibit substantial improvements over counterparts without Synth-SBDH training on real-world clinical datasets. For instance, Synth-SBDH yields performance gains of up to 63.75\% in SBDH detection as a multi-label classification task. Even in the more challenging named entity recognition task with a 2.3 times larger training dataset, synth-SBDH elevates the performance of sequence-to-sequence (Seq2seq) models.
    \item Synth-SBDH significantly improves the performance for rare SBDH categories on out-of-domain real-world clinical datasets, with up to 94.44 absolute F-score improvements. Synth-SBDH is also useful in low-resource settings. 
    \item To facilitate advancements in SBDH extraction and analysis, we make Synth-SBDH, its expert-annotated test set, and all relevant materials (including model checkpoints) publicly available. 
\end{enumerate}
With its comprehensive annotations and expert evaluation, Synth-SBDH can be a benchmark dataset for researchers and practitioners alike, offering insights into the intricate relationship between SBDH and health outcomes in EHR text.

\section{Related work}
\subsection{SBDH datasets}
Recognizing the significance of SBDH on various clinical outcomes and the utility of unstructured EHR notes, numerous studies have employed NLP techniques to extract SBDH information from EHR notes \citep{patra2021extracting,agnikula2021social}. Most previous studies on SBDH datasets have focused on individual SBDH categories, such as smoking \citep{uzuner2008identifying, savova2008mayo}, housing stability \citep{chapman2021rehoused, bejan2018mining}, substance abuse \citep{wang2015automated, alzoubi2018automated} etc. Conversely, \citet{yu2022assessing} developed a corpus of 15 SDOH categories from the EHR notes of lung cancer patients while \citet{mitra2023associations} built a corpus of 12 SBDH categories to evaluate their associations with suicide. \citet{han2022classifying} created a dataset of 13 SBDHs to demonstrate the effectiveness of deep-learning-based NLP systems for extracting SBDH from clinical text. However, none of these datasets are publicly available. The majority of existing publicly available SBDH datasets (Table \ref{dataset_comp}) are based on MIMIC-III, which requires additional credentials to access. In this work, we aim to provide a fully open-sourced SBDH dataset without compromising data privacy by eliminating the use of real-world EHR data. Our objective is to develop a more comprehensive SBDH dataset that includes a wider range of SBDH categories, encompassing both behavioral factors and annotation rationales. Furthermore, we investigate the utility of this dataset across multiple tasks on real-world SBDH datasets. 

\subsection{LLMs to generate data}
LLMs have been widely studied for their capacity to generate data for various general-domain tasks \citep{chung2023increasing,yoo2021gpt3mix,sahu2022data,hartvigsen2022toxigen}. \citet{chung2023increasing} investigated the advantages of human-AI collaboration to ensure high diversity and comprehensive coverage in data generated by LLMs. \citet{yoo2021gpt3mix} introduced GPT3mix, a straightforward text generation technique that uses few-shot examples from real-world data to prompt an LLM for synthetic data generation. \citet{sahu2022data} employed a similar prompt-based approach to generate data for intent classification (IC) and demonstrated that quality assurance, either through relabeling or filtering, is necessary for certain downstream tasks. \citet{hartvigsen2022toxigen} also adopted a few-shot prompting technique to create ToxiGen, a large-scale synthetic dataset consisting of both toxic and benign statements. In addition, they incorporated a novel adversarial classifier-in-the-loop decoding algorithm to produce a more challenging subset. 

With recent advancements in generative modeling, LLMs have also been explored within the clinical domain \citep{xu2023knowledge,li2023two, guo2023dr, chintagunta2021medically}. Although public EHR datasets such as MIMIC-III \citep{johnson2016mimic} and MIMIC-IV \citep{johnson2023mimic} exist, their access is limited to individuals with specific credentials, and any products developed using these datasets are subject to the same restrictions. This, combined with advancements in LLMs, has motivated researchers to leverage LLMs for generating clinical data, such as generating synthetic clinical notes \citep{kweon2023publicly} and patient-physician dialogues \citep{wang2024notechat} from case reports. Regarding SBDH, \citet{guevara2024large} generated a small dataset (n=480) and fine-tuned multiple LLMs to showcase the potential of such data. In contrast, this study introduces a significantly larger ($\sim$18x) synthetic SBDH dataset incorporating behavioral factors and annotation rationales.

\begin{figure*}[hbt]
    \centering
    \includegraphics[width=.9\linewidth,scale=1]{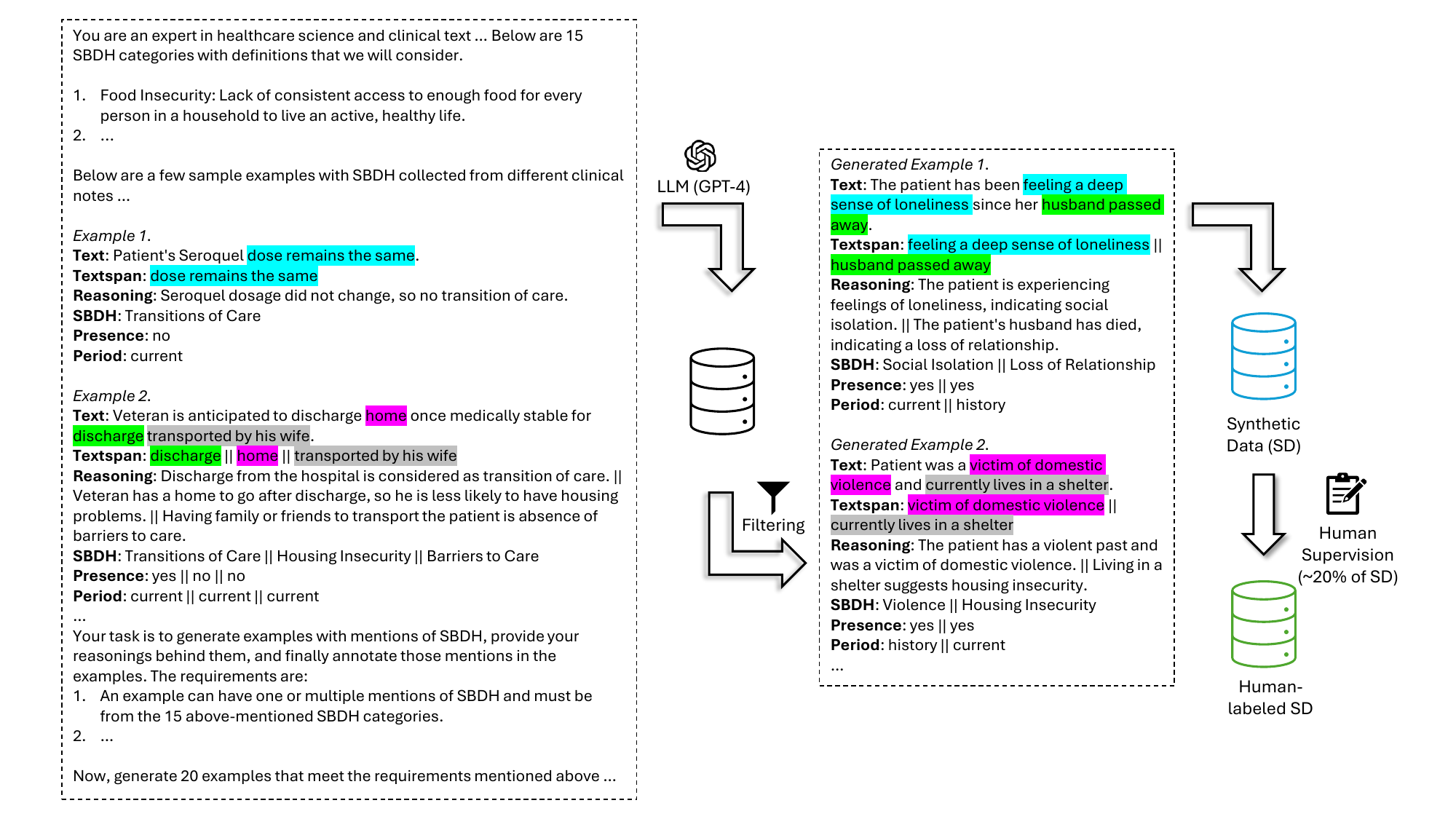}
    \caption{An illustration of the data generation process for Synth-SBDH.}
    \label{fig:data_gen}
    \vspace{-2mm}
\end{figure*}

\section{Synth-SBDH}
\subsection{Data generation}
The data generation process for Synth-SBDH consists of four stages as shown in Figure \ref{fig:data_gen}. The first stage starts with defining SBDH categories and seed examples. Based on expert opinion and a comprehensive literature review \citep{mitra2023associations,bossert2013measuring,nicholson2012review,daniel2018addressing,artiga2018beyond,SocialDe31:online,SocialDe61:online,Socialde81:online,USDAERSF54:online,hsrdrese29:online}, we selected 15 SBDH categories\footnote{This selection comprises 13 SBDH categories and two additional factors, `transition of care' and `pain'. Although neither of these is classified as an SBDH, experts identified them as relevant due to their frequent associations with various SBDH factors. For simplicity, we refer to this set collectively as 15 SBDH categories.} (Table \ref{synth_sbdh_ctg}). A group of two human experts provided three sample examples for each category, reflecting the typical language in clinical notes \footnote{Since data from different hospitals or even units within the same hospital often differ in language style \citep{yang2022machine}, we do not attempt to match any specific clinical cohort.}, resulting in a total of 45 seed examples. Each example includes a text sequence and all possible SBDH annotations encompassing attributes such as \textit{presence}, \textit{period}, and \textit{reasoning}. The \textit{presence} attribute (yes/no) indicates whether an SBDH is present. We include this attribute so that annotations with \textit{presence}=`no' can help models understand negations. The \textit{period} attribute provides temporal information indicating whether the SBDH described in the example is an event of the past (history) or present (current), while \textit{reasoning} offers a rationale behind each annotation. 

In the second stage, we prompt GPT-4 (GPT-4-0613, accessed via OpenAI API) with instructions to generate synthetic SBDH data based on the definitions and seed examples obtained from the first stage. This is an iterative process where at each iteration, we randomly sample 10 of the 45 seed examples and instruct the LLM to generate 20 synthetic examples with SBDH annotations and attributes. In total, this stage generated 14,244 examples with 22,459 annotations. The prompting template and seed examples are available in Appendix \ref{app_synth}.

\begin{table}
  \fontsize{6.5pt}{6.5pt}\selectfont
  \centering
  \begin{threeparttable}
  \begin{tabular}{llll}
    \toprule
         & Synth-SBDH     & MIMIC-SBDH$_\text{aligned}$   & VA-SBDH \\
    \midrule
    \#examples & 8,767  & 7,025        & 20,570 \\
    \#annotations & 14,342 & - & 46,703 \\
    \#categories & 15  & 4   & 12  \\
    Avg. seq. length\tnote{*} & 13.70 & 35.13 & 47.43 \\
    Avg. span length & 3.96 & - & 1.35 \\
    Public availability  & \cmark    & \cmark & \xmark\\
    \bottomrule
  \end{tabular}
  \begin{tablenotes}
    \item[*] In words.
  \end{tablenotes}
  \end{threeparttable}
  \caption{Dataset Statistics.}
  \label{exp_dataset_comp}
  \vspace{-5mm}
\end{table}

The third stage involves filtering the generated data to ensure output format and diversity. First, we filter out all examples that do not adhere to the expected output format according to the prompt (690 examples, 4.81\%). Next, we notice that GPT-4 tended to generate repetitive examples, even when prompted with different seed examples. So, following the methodology of \cite{wang2022self}, we use the ROUGE-L metric to set a similarity threshold for pairs of generated examples. An example is excluded from the final data pool if it has a ROUGE-L score of 0.7 or higher with all the previous examples in the pool (4,787 examples, 33.38\%). Our final data pool contains 8,767 examples with 14,342 annotations across 15 SBDH categories, constituting the Synth-SBDH dataset.

In the final stage, we randomly select 20\% of Synth-SBDH as the test set and conduct an evaluation by the same group of human experts from the first stage to provide an expert-supervised, high-quality test set that can be used as a gold-standard dataset in future studies. More about the human evaluation process and error analysis are detailed in sections \ref{human_eval} and \ref{err} respectively.

\subsection{Data statistics}
Tables \ref{dataset_comp} and \ref{exp_dataset_comp} provide key stats for Synth-SBDH. Synth-SBDH is the largest publicly available SBDH dataset covering a broad range of SBDH categories. With an average sequence length of 13.7 words and annotation span length of 3.96 words, Synth-SBDH examples exhibit mostly short sentences with long text spans for annotations. Definitions of all SBDH categories, example text spans, and their distributions are shown in Tables \ref{synth_sbdh_ctg} and \ref{tbl_synth_sbdh_ctg}. 
We make a 70:10:20 split to create training, development, and test sets. This results in 6,136, 836, and 1,755 examples respectively. For all experiments, we only use the training set to train a model, while the development set is retained for hyperparameter tuning. The test set is utilized for human evaluation and distillation experiments.

\section{Experiments}
\label{exp}

\subsection{Datasets}
We consider two real-world SBDH datasets to validate the utility of Synth-SBDH. Our first dataset is based on MIMIC-SBDH \citep{ahsan2021mimic}, a publicly available SBDH dataset. MIMIC-SBDH is curated from MIMIC-III notes and has 7 SBDH categories. To better align with the categories and definitions used in Synth-SBDH, we processed the 7 categories and created 4 SBDH categories. We refer to this dataset as MIMIC-SBDH$_\text{aligned}$. The second dataset is a subset of the SBDH dataset used by \citet{mitra2023associations}. This is a private\footnote{Access can be obtained by relevant approvals through US VA Office of Research and Development.} dataset based on EHR notes obtained from the US Veterans Health Administration (VHA) Corporate Data Warehouse. It contains annotations for 13 SBDH categories of which we excluded one category - `suicide outcome' as it was absent in Synth-SBDH. We refer to this dataset as VA-SBDH. Table \ref{exp_dataset_comp} shows summary statistics of the two datasets. Additional information about the datasets and how we process them are available in Appendix \ref{app_data}.

\subsection{Evaluation tasks}
\label{eval_tasks}
To assess the utility of Synth-SBDH, we select a set of carefully designed experiments to cover three aspects - \textbf{versatility}, \textbf{generalizability}, and \textbf{knowledge distillation} capability. For \textbf{versatility}, we want to see if Synth-SBDH can be applied to different SBDH detection tasks. In our experiments, we rephrase SBDH detection as two tasks - multi-label classification (MLC) (Figure \ref{fig:tasks}a) and named entity recognition (NER) (Figure \ref{fig:tasks}b). NER is more challenging than MLC, especially when the systems are evaluated based on exact matching (i.e. predicted/generated text span and gold span should be an exact match). However, NER enhances interpretability, as an NER model can provide the text span along with each predicted SBDH label, offering healthcare providers more insights into which section of the input sequence was annotated by the model. These experiments also help investigate the \textbf{generalizability} of Synth-SBDH, as we perform the tasks on two real-world datasets from different healthcare settings. For each task, we follow a three-step process - 1) process Synth-SBDH to fit the task description, 2) fine-tune models on the processed Synth-SBDH data, and 3) continue fine-tuning the same models on the task-specific real-world dataset. We compare these models with those that were only fine-tuned on task-specific real-world datasets to evaluate the advantage of our synthetic data. For MLC, we consider both MIMIC-SBDH$_\text{aligned}$ and VA-SBDH whereas for NER, we consider only VA-SBDH as MIMIC-SBDH$_\text{aligned}$ does not include token-level data. The processing steps for adapting Synth-SBDH to MLC and NER are available in Appendix \ref{app_task}. 

Furthermore, Synth-SBDH is the only SBDH dataset with a rationale for each SBDH annotation. So, we leverage the \textit{distilling step-by-step} (DSS) framework \citep{hsieh2023distilling} to assess if rationales generated by an LLM (GPT-4 in this case) can improve the performance of SBDH extraction in small language models (SLM). Specifically, DSS utilizes a multi-task learning approach, where a generative model is trained to simultaneously predict both class labels and corresponding rationales (Figure \ref{fig:tasks}c) for a downstream task. This is similar to \textbf{knowledge distillation}, where the LLM (GPT-4) acts as the teacher model (having generated the rationales) and an SLM serves as the student model. Here we reformulate SBDH extraction as a generative NER task instead of MLC to explore the rationales' importance in a more challenging situation. Given that no publicly available SBDH dataset with annotation rationales exists at the time of this work, we report our results on the expert-reviewed test set of Synth-SBDH. 

\subsection{Models and metrics}
In our study, we prioritize the use of SLMs to enable the deployment of such systems in remote healthcare centers, which often operate under limited computational resources. For MLC, we consider RoBERTa \citep{liu2019roberta}, clinicalRoBERTa \citep{lewis-etal-2020-pretrained}, Mamba \citep{gu2023mamba}, and ClinicalMamba \citep{yang2024clinicalmamba}. We try standard fine-tuning for encoder-only models and prompt-based fine-tuning \citep{gao2020making, ding2021prompt} for all models. For the NER task, in addition to encoder-only models such as RoBERTa and clinicalRoBERTa, we also include sequence-to-sequence (Seq2seq) models such as T5 \citep{radford2019language} and FLAN-T5 \citep{chung2022scaling}. We exclude mamba models for NER due to their poor performance in preliminary experiments.\footnote{We speculate this is due to the differences in the attention mechanisms between the encoder-only transformer architecture and state-space models in Mamba, although a more in-depth investigation is necessary for a conclusive remark.} DSS being a generative task, we choose only T5 and FLAN-T5 models. We consider the base variant for RoBERTa-series (RoBERTa-base-PM-M3-Voc-distill-align for clinicalRoBERTa) and T5-series models. To match the model size of RoBERTa models, we consider Mamba-130m and clincalMamba-130m. We report both micro and macro F scores for all experiments. For NER and DSS, we report scores following exact matching criteria. We also report p-values when comparing performance \citep{altman2011obtain}.
More training details and hyperparameter configuration for each model are available in Appendix \ref{app_hp}.

\section{Results}
\label{res}
\paragraph{MLC.} Table \ref{result_ml} shows the results for our MLC task on MIMIC-SBDH$_\text{aligned}$. Among the four models with two different fine-tuning strategies, the majority yield significantly better performance when fine-tuned on Synth-SBDH. For example, RoBERTa with standard fine-tuning achieves a 9.75\% (p-val<0.01) gain in macro F (59.27\% to 65.05\%) while with prompt-based fine-tuning there is an impressive 63.75\% (p-val<0.001) gain in macro F (55.64\% to 91.12\%). Similarly, clinicalMamba with prompt-based fine-tuning enjoys a 10.70\% (p-val<0.001) improvement in macro F (82.77\% to 91.63\%). Interestingly, Mamba, without any pre-training on MIMIC-III, significantly outperforms clinicalRoBERTa. It is worth noting that the increase in macro F scores is substantially higher than micro F, indicating the benefits of Synth-SBDH for rare categories. We elaborate on this in section \ref{abl}. Results for more MLC experiments are available in Appendix \ref{mlc_more_exp}.
\begin{table}[hbt]
  \fontsize{7.5pt}{9pt}\selectfont
  \centering
  \begin{threeparttable}
  \begin{tabular}{lcc}
    \toprule
    Model &Micro F &Macro F \\\midrule
    Standard Fine-tuning\\\midrule
    RoBERTa-base & 83.00 ± 2.01	 & 59.27 ± 3.22 \\
    \hspace{2mm} + With Synth-SBDH & \textbf{87.32 ± 0.86}  & \textbf{65.05 ± 0.90} \\
    \cdashline{1-3}
    ClinicalRoBERTa-base & 85.11 ± 0.48	& 61.14 ± 1.32  \\
    \hspace{2mm} + With Synth-SBDH & \textbf{89.32 ± 0.72}  & \textbf{66.90 ± 0.52} \\
    \midrule
    Prompt-based Fine-tuning\\ \midrule
     RoBERTa-base & 79.32 ± 0.92	& 55.64 ± 1.30  \\
    \hspace{2mm} + With Synth-SBDH & \textbf{91.11 ± 0.07}   & \textbf{91.11 ± 0.15}\\
    \cdashline{1-3}
    ClinicalRoBERTa-base & 86.08 ± 0.89	& 63.64 ± 1.50	    \\
    \hspace{2mm} + With Synth-SBDH & \textbf{91.41 ± 0.12}	 & \textbf{90.81 ± 0.73}\\
    \cdashline{1-3}
    Mamba-130m & 89.60 ± 0.33	& 85.44 ± 2.63 \\
    \hspace{2mm} + With Synth-SBDH & \textbf{90.32 ± 0.09}   & \textbf{89.89 ± 0.34} \\
    \cdashline{1-3}
    ClinicalMamba-130m  & 91.01 ± 0.70	& 82.77 ± 2.91 \\
    \hspace{2mm} + With Synth-SBDH & \textbf{91.89 ± 0.18}   & \textbf{91.63 ± 0.46}\\
    \bottomrule
  \end{tabular}
  \end{threeparttable}
  \caption{SBDH detection as an MLC task on MIMIC-SBDH$_\text{aligned}$. Fine-tuning baseline models on Synth-SBDH before fine-tuning on the target dataset yields performance improvements in most cases. Each cell value indicates the mean and standard deviation over three independent runs.}
  \label{result_ml}
\end{table} 

\begin{table}
  \fontsize{8pt}{10pt}\selectfont
  \centering
  \begin{threeparttable}
  \begin{tabular}{lcc}
    \toprule
    Model &Micro F &Macro F  \\\midrule
    Encoder-only models\\\midrule
    RoBERTa-base & 71.09 ± 0.11    & \textbf{67.99 ± 0.19}    \\
    \hspace{2mm} + With Synth-SBDH & \textbf{71.18 ± 0.12}    & 67.87 ± 0.25   \\
    \cdashline{1-3}
    ClinicalRoBERTa-base & 69.95 ± 0.12   & 66.03 ± 0.27  \\
    \hspace{2mm} + With Synth-SBDH & \textbf{70.33 ± 0.11}    & \textbf{66.33 ± 0.39}  \\
    \midrule
    Seq2Seq models\\ \midrule
    T5-base & 28.28 ± 19.93    & 25.44 ± 18.24   \\
    \hspace{2mm} + With Synth-SBDH & \textbf{64.70 ± 0.77}    & \textbf{60.73 ± 0.94}    \\
    \cdashline{1-3}
    FLAN-T5-base & 68.77 ± 0.11    & 64.93 ± 0.13  \\
    \hspace{2mm} + With Synth-SBDH & \textbf{69.59 ± 0.23}    & \textbf{66.34 ± 0.31}   \\
    \bottomrule
  \end{tabular}
  \end{threeparttable}
  \caption{SBDH detection as an NER task on VA-SBDH. Fine-tuning baseline models on Synth-SBDH before fine-tuning on the target dataset improves performance in almost all cases. Each cell value indicates the mean and standard deviation over three independent runs. Results with relaxed matching are available in Table \ref{result_ner_relaxed}.}
  \label{result_ner}
  \vspace{-4mm}
\end{table}

\paragraph{NER.} For the NER task, we also observe an increase in F scores across most models with Synth-SBDH, as shown in Table \ref{result_ner}. We notice the most gains for Seq2Seq models - 138.72\%\footnote{We found T5-base to be highly unstable on VA-SBDH for NER. In contrast, T5-base with Synth-SBDH showed relatively stable behavior. FLAN-T5-base does not show such behavior - possibly its additional instruction-tuning contributes to greater training stability and better convergence behavior.} (p-val<0.001) for T5 and 2.17\% (p-val<0.001) FLAN-T5. However, gains for encoder-only models are not as statistically significant as in the MLC task. This can be attributed to two factors - 1) VA-SBDH is more balanced than MIMIC-SBDH$_\text{aligned}$ (Tables \ref{tbl_mimic_sbdh_ctg} and \ref{tbl_va_sbdh_ctg}), so, even categories with low prevalence have enough examples in the dataset and thus, decreases any reliance on Synth-SBDH and 2) VA-SBDH is almost 2.3 times bigger than Synth-SBDH and sourced from real clinical notes with human annotations, therefore, it limits the potential benefits of a much smaller synthetic dataset. Despite these challenges, we notice that Synth-SBDH can make less performant models more competitive. For example, fine-tuning on Synth-SBDH made T5 more stable. This highlights the potential benefits of synthetic data for an unstable training pipeline.

\begin{table}
  \fontsize{8pt}{8pt}\selectfont
  \centering
  \begin{threeparttable}
  \begin{tabular}{lccc}
    \toprule
    Model &Micro F &Macro F \\\midrule
    T5-base \\
    \hspace{2mm} Standard Fine-Tuning & 57.65 ± 0.30    & 54.94 ± 0.02   \\
    \hspace{2mm} DSS & \textbf{58.36 ± 0.26}    & \textbf{55.60 ± 0.26}  \\
    \cdashline{1-4}
    FLAN-T5-base \\ 
    \hspace{2mm} Standard Fine-Tuning & 57.19 ± 0.44    & 54.36 ± 0.50 \\
    \hspace{2mm} DSS & \textbf{57.70 ± 0.89}    & \textbf{54.94 ± 0.84} \\
    \bottomrule
  \end{tabular}
  \end{threeparttable}
  \caption{SBDH extraction with DSS on Synth-SBDH. All models with DSS outperformed standard fine-tuning. Models are evaluated on the expert-reviewed test set. Each cell value indicates the mean and standard deviation over three independent runs.}
  \label{result_dss}
  \vspace{-5mm}
\end{table}

\paragraph{DSS.} We show the results in Table \ref{result_dss}. Our findings suggest that the DSS framework can improve the overall performance of small generative models on generative SBDH extraction task. In our experiments, both T5 and FLAN-T5 achieve 1.20\% (p-val=0.01) and 1.07\% (p-val=0.36) increase in macro F scores respectively. Note that the original framework utilized real-world data to generate rationales by an LLM whereas Synth-SBDH rationales were generated during the data generation process. Our pipeline also differs from the original work in that they considered natural language inference, question answering, and mathematical problem-solving tasks whereas we considered NER, a previously uncharted territory for the DSS framework.

\begin{table*}[hbt!]
  \fontsize{8pt}{10pt}\selectfont
  \centering
  \begin{threeparttable}
  \begin{tabular}{lcccc}
    \toprule
    Model & \makecell{Substance\\abuse\\(n = 547)} & \makecell{Housing\\insecurity\\(n = 18)} & \makecell{Financial\\insecurity\\(n = 378)} & \makecell{Isolation or loss \\of relationship\\(n = 154)}\\\midrule
    RoBERTa-base & 87.10 ± 0.13	& 0.00 ± 0.00	& 79.66 ± 0.84	& 55.79 ± 4.36  \\
    \hspace{2mm} + Fine-tuned on Synth-SBDH & \textbf{90.91 ± 0.12}	& \textbf{94.44 ± 0.00}	& \textbf{93.44 ± 0.04}	& \textbf{85.64 ± 0.66}\\
    \cdashline{1-5}
    ClinicalRoBERTa-base & 89.35 ± 0.29	& 3.33 ± 4.71	& 87.89 ± 1.73	& 74.00 ± 0.95	    \\
    \hspace{2mm} + Fine-tuned on Synth-SBDH & \textbf{91.07 ± 0.26}    & \textbf{90.74 ± 2.62}   & \textbf{92.52 ± 0.13}	 & \textbf{88.89 ± 0.32}\\
    \cdashline{1-5}
    Mamba-130m & 89.71 ± 0.55	& 75.14 ± 9.74	& 91.71 ± 0.53	& 85.20 ± 0.93 \\
    \hspace{2mm} + Fine-tuned on Synth-SBDH & \textbf{89.65 ± 0.60}	& \textbf{90.63 ± 2.46}   & \textbf{92.80 ± 0.40}   & \textbf{86.48 ± 2.41} \\
    \cdashline{1-5}
    ClinicalMamba-130m & \textbf{91.76 ± 0.25}	& 60.84 ± 8.57	& \textbf{93.11 ± 0.70}	& 85.39 ± 2.35 \\
    \hspace{2mm} + Fine-tuned on Synth-SBDH & 91.54 ± 0.27	& \textbf{94.34 ± 0.15}	& 92.85 ± 0.43	& \textbf{87.81 ± 1.31}\\
    \bottomrule
  \end{tabular}
  \end{threeparttable}
  \caption{Change of F-scores across the 4 SBDH categories on MIMIC-SBDH$_\text{aligned}$ after fine-tuning on Synth-SBDH. Rare categories such as `Housing insecurity' and `Isolation or loss of relationship' benefited the most from Synth-SBDH. All results are with prompt-based fine-tuning, each cell value indicates the mean and standard deviation over three independent runs.}
  \label{result_rare}
  \vspace{-7mm}
\end{table*}

\section{Human and LLM evaluation}
\label{human_eval}
Given that Synth-SBDH is a synthetic dataset, we conducted a comprehensive human evaluation to assess its quality. We selected the entire test set (1,755 examples) for this purpose. The same group of two human experts who provided seed examples in the synthetic data generation phase assessed all test annotations and the quality of the provided rationales. Experts considered each GPT-4 generated annotation for one of these actions - \textit{keep} it when the annotation is correct, \textit{update} it when there is an error, or \textit{discard} it if the annotation was unnecessary or erroneous. Additionally, the experts were instructed to \textit{add} missing annotations. Synth-SBDH obtained an LLM-Human agreement of 71.06\% while the two experts had an inter-annotator agreement of 77.80\%\footnote{All agreements are in percentage agreement and measured across four aspects - text span, SBDH, presence, and period. We adopted relaxed matching for text span and strict matching for SBDH, presence, and period.}. For annotation disagreements between the experts, multiple adjudication sessions were carried out until convergence and evaluation guidelines were updated accordingly. We would like to note that the LLM-Human agreement of 71.06\% is not far off from the inter-annotator agreement of 77.80\%. The latter indicates the challenges and subjectivity of annotating SBDH categories, even for experts. Additionally, studies have demonstrated that synthetic data with high label noise \citep{schick-schutze-2021-generating} or low human-AI alignment \citep{wang2022self} can still be beneficial in various downstream tasks.

The experts also rated the GPT-4-generated rationales on a 4-point Likert scale, yielding an aggregated rating of 3.66. Based on the expert-reviewed annotations, we created a silver-level test set, which we also release with Synth-SBDH. Furthermore, we curated a list of notable trends observed from successful and failed annotations in Synth-SBDH, available in Appendix \ref{eval_obs}. Similar to human evaluation, we also conducted LLM evaluation by prompting GPT-4, resulting in a 90.78\% LLM-LLM alignment. All evaluation guidelines are provided in Appendix \ref{eval}.

\section{Discussions}
\subsection{Ablations}
\label{abl}
\paragraph{Synth-SBDH benefits rare categories.} In section \ref{res}, we showed that fine-tuning on Synth-SBDH improves macro F scores more substantially than micro F scores on MIMIC-SBDH$_\text{aligned}$, leading us to hypothesize that rare SBDH categories benefit from synthetic data of relevant categories from Synth-SBDH. To validate this, we compare the F scores of all SBDH categories side-by-side for models with and without Synth-SBDH fine-tuning as shown in Table \ref{result_rare}. This clearly shows that for SBDH categories with low prevalence such as `Housing insecurity' (n=18) and `Isolation or loss of relationship' (n=154), fine-tuning on Synth-SBDH, indeed, yielded the most improvements compared to the other categories. 

\begin{figure}[hbt!]
    \centering
    \includegraphics[width=.8\linewidth]{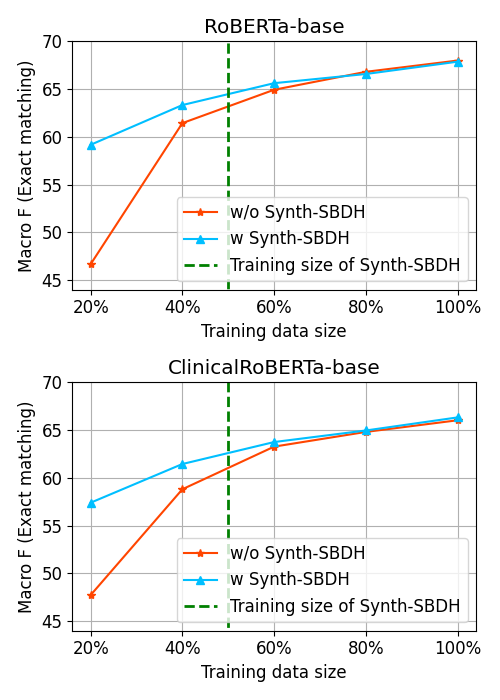}
    \caption{Impact of training size. Performance gain diminishes as the training size of VA-SBDH surpasses that of Synth-SBDH}
    \label{fig:perf_size}
    \vspace{-6mm}
\end{figure}

\paragraph{Performance with data size.} We vary the training data size of VA-SBDH to investigate how a change in data size interacts with fine-tuning on Synth-SBDH. 
Figure \ref{fig:perf_size} shows that Synth-SBDH helps the most in the low training data regime, as expected. Gain diminishes as the training size surpasses that of Synth-SBDH. 
Interestingly, the performance curves have not plateaued for any of the settings, indicating room for improvement with more data. As annotating actual EHR notes is time-consuming, expensive, and raises privacy concerns, developing an even larger synthetic SBDH dataset following a robust framework holds great potential.

\subsection{Error analysis}
\label{err}
We conducted a qualitative analysis of 2,906 SBDH annotations from 1,755 examples (Synth-SBDH test set). As outlined in section \ref{human_eval}, experts were instructed to carry out one of four operations - \textit{Keep}, \textit{Update}, \textit{Discard}, and \textit{Add}. A detailed category-wise breakdown is presented in Table \ref{sbdh_ctg_eval}. Our analysis reveals that human experts agreed with 2,478  GPT-4 generated annotations (71.06\%, \textit{Keep}). Additionally, experts added 581 new annotations (16.66\%, \textit{Add}), increasing the total annotation count to 3,487. After accounting for the 142 discarded annotations, the final annotation count stands at 3,345 from 1,732 examples. The categories `Physical Isolation', `Loss of Relationship', and `Violence' posed the greatest challenge for GPT-4 to annotate, whereas `Pain', `Food Insecurity', and `Job Insecurity', were the easiest. Furthermore, among the missed annotations, `Loss of Relationship', `Housing Insecurity', and `Social Isolation' were the most frequent categories.

\section{Conclusion}
In this study, we present and release Synth-SBDH, a synthetic dataset for SBDH in clinical text. Surpassing existing publicly available SBDH datasets in size, Synth-SBDH encompasses a diverse array of SBDH categories along with relevant information such as presence and annotation rationales, thereby addressing critical gaps in the existing datasets. Our extensive evaluations demonstrate that models trained on Synth-SBDH achieve significant performance improvements when applied to real-world SBDH datasets from two distinct hospital settings. Additionally, the inclusion of SBDH rationales aids in distilling reasoning capability into smaller models, enhancing their SBDH detection. Synth-SBDH proves to be highly beneficial for identifying rare SBDH categories and for developing systems within low-resource settings. Synth-SBDH not only highlights the potential of synthetic data in mitigating data scarcity and privacy but also advances more effective and inclusive healthcare analytics.

\section*{Limitations and future work}
\label{limit}
Our study has several limitations. First, our seed examples and SBDH definitions are based on VHA EHR notes, and they do not fully capture the diverse clinical text styles found in typical notes, such as medical abbreviations, redactions, and bullet points. Creating examples and definitions using different hospitals' EHR data banks as well as varying text formats could enhance its diversity and generalizability. This will be considered in the next version of Synth-SBDH. Second, in all of our experiments, we avoided the use of LLMs. However, this was a conscious choice to emulate real-world scenarios at remote healthcare centers with low computational resources and limited data availability. Nevertheless, we recognize the potential of LLMs and report the performance of LLMs such as Llama 3.2 3B \citep{llama3_2modelcard}, Llama 3 8B \citep{llama3modelcard}, and FLAN-T5-xl \citep{chung2022scaling} in Appendix \ref{llm}. Third, Synth-SBDH is relatively small compared to other domain-specific synthetic datasets. However, Synth-SBDH can still yield performance gain on a 2.3x bigger, balanced, and more expensive\footnote{Creating VA-SBDH cost approximately \$413k (~\$8.84/annotation) and took over 2.5 years, excluding costs related to training annotators, server hosting, and IT support for the annotation process. In contrast, Synth-SBDH was generated quickly (within days) at a cost of approximately \$90 (\$0.006/annotation).} real-world dataset, highlighting the potential Synth-SBDH holds if its size is increased. We leave this for future work. Fourth, Synth-SBDH uses a very simple data generation framework. Techniques such as self-consistency \citep{wang2022self}, multi-agent framework \citep{hong2023metagpt}, self-verification \citep{weng2022large} etc. may improve the quality of synthetic data and this warrants future exploration. Fifth, Synth-SBDH only contains examples in the English language. Exploring the impact of synthetic data for SBDH detection in other languages is a promising future research direction. Finally, while MIMIC-III is accessible only through PhysioNet by credentialed users, verifying potential data leakage between GPT-4’s pretraining data and MIMIC-III (or other open-source datasets) is nearly impossible, and it is difficult to guarantee no data leakage. Furthermore, LLMs like GPT-4 may introduce biases in clinical applications \cite{zack2024assessing,omiye2023large,omar2025evaluating}. Although this study did not focus on bias detection, future research should explore biases in synthetic SBDH data and their implications for health outcomes to support the development of more equitable healthcare systems.

\section*{Societal Impacts}
\label{soc_impact}
Synth-SBDH has the potential to significantly enhance healthcare research and practice by providing a robust and comprehensive resource for the extraction and analysis of SBDH. By offering a large, publicly available synthetic dataset, Synth-SBDH enables researchers to develop and refine NLP models without the privacy concerns associated with real clinical data. This accessibility can lead to accelerated advancements in understanding the nuanced interplay between SBDH and health outcomes, thereby improving patient care and population health management. Moreover, the dataset's versatility and generalizability can enhance the detection and analysis of SBDH across diverse clinical scenarios, contributing to more effective and equitable healthcare interventions. In general, synthetic clinical data generated under careful ethical considerations and responsible guidelines has the potential to address biases found in real-world clinical data \citep{chen2021synthetic,yoon2023ehr}.

However, the deployment of synthetic datasets like Synth-SBDH also raises potential concerns. The reliance on synthetic data, while addressing privacy issues, may introduce biases or inaccuracies that do not fully capture the complexities of real-world data. If not carefully validated, models trained on such data could propagate these biases, potentially leading to erroneous conclusions or suboptimal clinical decisions. For example, LLMs are known to have cognitive \citep{schmidgall2024addressing} and racial biases \citep{omiye2023large}. Ethical concerns such as lack of human empathy and trust, authorship, and privacy of input text are also relevant in using LLM-generated data \citep{li2023ethics}. Additionally, the use of LLMs in generating this dataset underscores the dependency on advanced and intensive computational resources, which may not be equally accessible to all research institutions, thereby perpetuating disparities in research capabilities. 


\bibliography{acl_latex}

\begin{thebibliography}{90}
\providecommand{\natexlab}[1]{#1}

\bibitem[{2030()}]{SocialDe61:online}
Healthy~People 2030.
\newblock Social determinants of health.
\newblock \url{https://health.gov/healthypeople/priority-areas/social-determinants-health}.
\newblock (Accessed on 08/16/2024).

\bibitem[{Achiam et~al.(2023)Achiam, Adler, Agarwal, Ahmad, Akkaya, Aleman, Almeida, Altenschmidt, Altman, Anadkat et~al.}]{achiam2023gpt}
Josh Achiam, Steven Adler, Sandhini Agarwal, Lama Ahmad, Ilge Akkaya, Florencia~Leoni Aleman, Diogo Almeida, Janko Altenschmidt, Sam Altman, Shyamal Anadkat, et~al. 2023.
\newblock Gpt-4 technical report.
\newblock \emph{arXiv preprint arXiv:2303.08774}.

\bibitem[{Adkins-Jackson et~al.(2023)Adkins-Jackson, George, Besser, Hyun, Lamar, Hill-Jarrett, Bubu, Flatt, Heyn, Cicero et~al.}]{adkins2023structural}
Paris~B Adkins-Jackson, Kristen~M George, Lilah~M Besser, Jinshil Hyun, Melissa Lamar, Tanisha~G Hill-Jarrett, Omonigho~M Bubu, Jason~D Flatt, Patricia~C Heyn, Ethan~C Cicero, et~al. 2023.
\newblock The structural and social determinants of alzheimer's disease related dementias.
\newblock \emph{Alzheimer's \& Dementia}, 19(7):3171--3185.

\bibitem[{Agnikula et~al.(2021)Agnikula, Balls-BerryJoyce~Joy et~al.}]{agnikula2021social}
KshatriyaBhavani~Singh Agnikula, E~Balls-BerryJoyce~Joy, et~al. 2021.
\newblock Social and behavioral determinants of health in the era of artificial intelligence with electronic health records: a scoping review.
\newblock \emph{Health Data Science}.

\bibitem[{Ahsan et~al.(2021)Ahsan, Ohnuki, Mitra, and Yu}]{ahsan2021mimic}
Hiba Ahsan, Emmie Ohnuki, Avijit Mitra, and Hong Yu. 2021.
\newblock Mimic-sbdh: a dataset for social and behavioral determinants of health.
\newblock In \emph{Machine Learning for Healthcare Conference}, pages 391--413. PMLR.

\bibitem[{AI@Meta(2024{\natexlab{a}})}]{llama3modelcard}
AI@Meta. 2024{\natexlab{a}}.
\newblock \href {https://github.com/meta-llama/llama-models/blob/main/models/llama3/MODEL_CARD.md} {Llama 3 model card}.

\bibitem[{AI@Meta(2024{\natexlab{b}})}]{llama3_2modelcard}
AI@Meta. 2024{\natexlab{b}}.
\newblock \href {https://github.com/meta-llama/llama-models/blob/main/models/llama3_2/MODEL_CARD.md} {Llama 3.2 model card}.

\bibitem[{Altman and Bland(2011)}]{altman2011obtain}
Douglas~G Altman and J~Martin Bland. 2011.
\newblock How to obtain the p value from a confidence interval.
\newblock \emph{Bmj}, 343.

\bibitem[{Alzoubi et~al.(2018)Alzoubi, Ramzan, Alzubi, and Mesbahi}]{alzoubi2018automated}
Hadeel Alzoubi, Naeem Ramzan, Raid Alzubi, and Ehsan Mesbahi. 2018.
\newblock An automated system for identifying alcohol use status from clinical text.
\newblock In \emph{2018 International Conference on Computing, Electronics \& Communications Engineering (iCCECE)}, pages 41--46. IEEE.

\bibitem[{Artiga et~al.(2018)Artiga, Hinton et~al.}]{artiga2018beyond}
Samantha Artiga, Elizabeth Hinton, et~al. 2018.
\newblock Beyond health care: the role of social determinants in promoting health and health equity.
\newblock \emph{Kaiser Family Foundation}, 10.

\bibitem[{based Synthesis Program (ESP)~Center()}]{hsrdrese29:online}
Evidence based Synthesis Program (ESP)~Center.
\newblock Evidence review: Social determinants of health for veterans.
\newblock \url{https://www.hsrd.research.va.gov/publications/esp/socialdeterminants-REPORT.pdf}.
\newblock (Accessed on 08/16/2024).

\bibitem[{Bejan et~al.(2018)Bejan, Angiolillo, Conway, Nash, Shirey-Rice, Lipworth, Cronin, Pulley, Kripalani, Barkin et~al.}]{bejan2018mining}
Cosmin~A Bejan, John Angiolillo, Douglas Conway, Robertson Nash, Jana~K Shirey-Rice, Loren Lipworth, Robert~M Cronin, Jill Pulley, Sunil Kripalani, Shari Barkin, et~al. 2018.
\newblock Mining 100 million notes to find homelessness and adverse childhood experiences: 2 case studies of rare and severe social determinants of health in electronic health records.
\newblock \emph{Journal of the American Medical Informatics Association}, 25(1):61--71.

\bibitem[{Blosnich et~al.(2020)Blosnich, Montgomery, Dichter, Gordon, Kavalieratos, Taylor, Ketterer, and Bossarte}]{blosnich2020social}
John~R Blosnich, Ann~Elizabeth Montgomery, Melissa~E Dichter, Adam~J Gordon, Dio Kavalieratos, Laura Taylor, Bryan Ketterer, and Robert~M Bossarte. 2020.
\newblock Social determinants and military veterans’ suicide ideation and attempt: a cross-sectional analysis of electronic health record data.
\newblock \emph{Journal of general internal medicine}, 35:1759--1767.

\bibitem[{Bossert and D'AMBROSIO(2013)}]{bossert2013measuring}
Walter Bossert and CONCHITA D'AMBROSIO. 2013.
\newblock Measuring economic insecurity.
\newblock \emph{International Economic Review}, 54(3):1017--1030.

\bibitem[{Canada.ca()}]{Socialde81:online}
Canada.ca.
\newblock Social determinants of health and health inequalities.
\newblock \url{https://www.canada.ca/en/public-health/services/health-promotion/population-health/what-determines-health.html}.
\newblock (Accessed on 08/16/2024).

\bibitem[{CDC()}]{SocialDe31:online}
CDC.
\newblock Social determinants of health (sdoh).
\newblock \url{https://www.cdc.gov/about/priorities/why-is-addressing-sdoh-important.html}.
\newblock (Accessed on 08/16/2024).

\bibitem[{Chapman et~al.(2021)Chapman, Jones, Kelley, Jones, Gawron, Montgomery, Byrne, Suo, Cook, Pettey et~al.}]{chapman2021rehoused}
Alec~B Chapman, Audrey Jones, A~Taylor Kelley, Barbara Jones, Lori Gawron, Ann~Elizabeth Montgomery, Thomas Byrne, Ying Suo, James Cook, Warren Pettey, et~al. 2021.
\newblock Rehoused: A novel measurement of veteran housing stability using natural language processing.
\newblock \emph{Journal of biomedical informatics}, 122:103903.

\bibitem[{Chen et~al.(2020)Chen, Tan, and Padman}]{chen2020social}
Min Chen, Xuan Tan, and Rema Padman. 2020.
\newblock Social determinants of health in electronic health records and their impact on analysis and risk prediction: a systematic review.
\newblock \emph{Journal of the American Medical Informatics Association}, 27(11):1764--1773.

\bibitem[{Chen et~al.(2021)Chen, Lu, Chen, Williamson, and Mahmood}]{chen2021synthetic}
Richard~J Chen, Ming~Y Lu, Tiffany~Y Chen, Drew~FK Williamson, and Faisal Mahmood. 2021.
\newblock Synthetic data in machine learning for medicine and healthcare.
\newblock \emph{Nature Biomedical Engineering}, 5(6):493--497.

\bibitem[{Chintagunta et~al.(2021)Chintagunta, Katariya, Amatriain, and Kannan}]{chintagunta2021medically}
Bharath Chintagunta, Namit Katariya, Xavier Amatriain, and Anitha Kannan. 2021.
\newblock Medically aware gpt-3 as a data generator for medical dialogue summarization.
\newblock In \emph{Machine Learning for Healthcare Conference}, pages 354--372. PMLR.

\bibitem[{Chung et~al.(2022)Chung, Hou, Longpre, Zoph, Tay, Fedus, Li, Wang, Dehghani, Brahma et~al.}]{chung2022scaling}
Hyung~Won Chung, Le~Hou, Shayne Longpre, Barret Zoph, Yi~Tay, William Fedus, Yunxuan Li, Xuezhi Wang, Mostafa Dehghani, Siddhartha Brahma, et~al. 2022.
\newblock Scaling instruction-finetuned language models.
\newblock \emph{arXiv preprint arXiv:2210.11416}.

\bibitem[{Chung et~al.(2023)Chung, Kamar, and Amershi}]{chung2023increasing}
John Joon~Young Chung, Ece Kamar, and Saleema Amershi. 2023.
\newblock Increasing diversity while maintaining accuracy: Text data generation with large language models and human interventions.
\newblock \emph{arXiv preprint arXiv:2306.04140}.

\bibitem[{Daniel et~al.(2018)Daniel, Bornstein, Kane, Health, and of~the American College~of Physicians*}]{daniel2018addressing}
Hilary Daniel, Sue~S Bornstein, Gregory~C Kane, Health, and Public Policy~Committee of~the American College~of Physicians*. 2018.
\newblock Addressing social determinants to improve patient care and promote health equity: an american college of physicians position paper.
\newblock \emph{Annals of internal medicine}, 168(8):577--578.

\bibitem[{Dasgupta et~al.(2018)Dasgupta, Beletsky, and Ciccarone}]{dasgupta2018opioid}
Nabarun Dasgupta, Leo Beletsky, and Daniel Ciccarone. 2018.
\newblock Opioid crisis: no easy fix to its social and economic determinants.
\newblock \emph{American journal of public health}, 108(2):182--186.

\bibitem[{Del{\'e}tang et~al.(2023)Del{\'e}tang, Ruoss, Duquenne, Catt, Genewein, Mattern, Grau-Moya, Wenliang, Aitchison, Orseau et~al.}]{deletang2023language}
Gr{\'e}goire Del{\'e}tang, Anian Ruoss, Paul-Ambroise Duquenne, Elliot Catt, Tim Genewein, Christopher Mattern, Jordi Grau-Moya, Li~Kevin Wenliang, Matthew Aitchison, Laurent Orseau, et~al. 2023.
\newblock Language modeling is compression.
\newblock \emph{arXiv preprint arXiv:2309.10668}.

\bibitem[{Dettmers et~al.(2023)Dettmers, Pagnoni, Holtzman, and Zettlemoyer}]{dettmers2023qlora}
Tim Dettmers, Artidoro Pagnoni, Ari Holtzman, and Luke Zettlemoyer. 2023.
\newblock Qlora: efficient finetuning of quantized llms (2023).
\newblock \emph{arXiv preprint arXiv:2305.14314}, 52:3982--3992.

\bibitem[{Ding et~al.(2021)Ding, Chen, Han, Xu, Xie, Zheng, Liu, Li, and Kim}]{ding2021prompt}
Ning Ding, Yulin Chen, Xu~Han, Guangwei Xu, Pengjun Xie, Hai-Tao Zheng, Zhiyuan Liu, Juanzi Li, and Hong-Gee Kim. 2021.
\newblock Prompt-learning for fine-grained entity typing.
\newblock \emph{arXiv preprint arXiv:2108.10604}.

\bibitem[{Dorr et~al.(2019)Dorr, Bejan, Pizzimenti, Singh, Storer, and Quinones}]{dorr2019identifying}
David Dorr, Cosmin~A Bejan, Christie Pizzimenti, Sumeet Singh, Matt Storer, and Ana Quinones. 2019.
\newblock Identifying patients with significant problems related to social determinants of health with natural language processing.
\newblock In \emph{MEDINFO 2019: Health and Wellbeing e-Networks for All}, pages 1456--1457. IOS Press.

\bibitem[{Franke(2014)}]{franke2014toxic}
Hillary~A Franke. 2014.
\newblock Toxic stress: effects, prevention and treatment.
\newblock \emph{Children}, 1(3):390--402.

\bibitem[{Fu et~al.(2024)Fu, Ramachandran, Dobbins, Park, Leu, Rosenberg, Lybarger, Xia, Uzuner, and Yetisgen}]{fu2024extracting}
Yujuan Fu, Giridhar~Kaushik Ramachandran, Nicholas~J Dobbins, Namu Park, Michael Leu, Abby~R Rosenberg, Kevin Lybarger, Fei Xia, Ozlem Uzuner, and Meliha Yetisgen. 2024.
\newblock Extracting social determinants of health from pediatric patient notes using large language models: Novel corpus and methods.
\newblock \emph{arXiv preprint arXiv:2404.00826}.

\bibitem[{Gao et~al.(2020)Gao, Fisch, and Chen}]{gao2020making}
Tianyu Gao, Adam Fisch, and Danqi Chen. 2020.
\newblock Making pre-trained language models better few-shot learners.
\newblock \emph{arXiv preprint arXiv:2012.15723}.

\bibitem[{Gu and Dao(2023)}]{gu2023mamba}
Albert Gu and Tri Dao. 2023.
\newblock Mamba: Linear-time sequence modeling with selective state spaces.
\newblock \emph{arXiv preprint arXiv:2312.00752}.

\bibitem[{Guevara et~al.(2024)Guevara, Chen, Thomas, Chaunzwa, Franco, Kann, Moningi, Qian, Goldstein, Harper et~al.}]{guevara2024large}
Marco Guevara, Shan Chen, Spencer Thomas, Tafadzwa~L Chaunzwa, Idalid Franco, Benjamin~H Kann, Shalini Moningi, Jack~M Qian, Madeleine Goldstein, Susan Harper, et~al. 2024.
\newblock Large language models to identify social determinants of health in electronic health records.
\newblock \emph{NPJ digital medicine}, 7(1):6.

\bibitem[{Guo et~al.(2023)Guo, Wang, Wang, and Yu}]{guo2023dr}
Zhen Guo, Peiqi Wang, Yanwei Wang, and Shangdi Yu. 2023.
\newblock Dr. llama: Improving small language models on pubmedqa via generative data augmentation.
\newblock \emph{arXiv preprint arXiv:2305.07804}.

\bibitem[{Han et~al.(2022)Han, Zhang, Shi, Richie, Liu, Tseng, Quan, Ryan, Brent, and Tsui}]{han2022classifying}
Sifei Han, Robert~F Zhang, Lingyun Shi, Russell Richie, Haixia Liu, Andrew Tseng, Wei Quan, Neal Ryan, David Brent, and Fuchiang~R Tsui. 2022.
\newblock Classifying social determinants of health from unstructured electronic health records using deep learning-based natural language processing.
\newblock \emph{Journal of biomedical informatics}, 127:103984.

\bibitem[{Hartvigsen et~al.(2022)Hartvigsen, Gabriel, Palangi, Sap, Ray, and Kamar}]{hartvigsen2022toxigen}
Thomas Hartvigsen, Saadia Gabriel, Hamid Palangi, Maarten Sap, Dipankar Ray, and Ece Kamar. 2022.
\newblock Toxigen: A large-scale machine-generated dataset for adversarial and implicit hate speech detection.
\newblock \emph{arXiv preprint arXiv:2203.09509}.

\bibitem[{Heidari et~al.(2023)Heidari, Zalmai, Richards, Sakthisivabalan, and Brown}]{heidari2023z}
Elham Heidari, Rana Zalmai, Kristin Richards, Lakshya Sakthisivabalan, and Carolyn Brown. 2023.
\newblock Z-code documentation to identify social determinants of health among medicaid beneficiaries.
\newblock \emph{Research in Social and Administrative Pharmacy}, 19(1):180--183.

\bibitem[{Hong et~al.(2023)Hong, Zheng, Chen, Cheng, Wang, Zhang, Wang, Yau, Lin, Zhou et~al.}]{hong2023metagpt}
Sirui Hong, Xiawu Zheng, Jonathan Chen, Yuheng Cheng, Jinlin Wang, Ceyao Zhang, Zili Wang, Steven Ka~Shing Yau, Zijuan Lin, Liyang Zhou, et~al. 2023.
\newblock Metagpt: Meta programming for multi-agent collaborative framework.
\newblock \emph{arXiv preprint arXiv:2308.00352}.

\bibitem[{Hsieh et~al.(2023)Hsieh, Li, Yeh, Nakhost, Fujii, Ratner, Krishna, Lee, and Pfister}]{hsieh2023distilling}
Cheng-Yu Hsieh, Chun-Liang Li, Chih-Kuan Yeh, Hootan Nakhost, Yasuhisa Fujii, Alexander Ratner, Ranjay Krishna, Chen-Yu Lee, and Tomas Pfister. 2023.
\newblock Distilling step-by-step! outperforming larger language models with less training data and smaller model sizes.
\newblock \emph{arXiv preprint arXiv:2305.02301}.

\bibitem[{Johnson et~al.(2023)Johnson, Bulgarelli, Shen, Gayles, Shammout, Horng, Pollard, Hao, Moody, Gow et~al.}]{johnson2023mimic}
Alistair~EW Johnson, Lucas Bulgarelli, Lu~Shen, Alvin Gayles, Ayad Shammout, Steven Horng, Tom~J Pollard, Sicheng Hao, Benjamin Moody, Brian Gow, et~al. 2023.
\newblock Mimic-iv, a freely accessible electronic health record dataset.
\newblock \emph{Scientific data}, 10(1):1.

\bibitem[{Johnson et~al.(2016)Johnson, Pollard, Shen, Lehman, Feng, Ghassemi, Moody, Szolovits, Anthony~Celi, and Mark}]{johnson2016mimic}
Alistair~EW Johnson, Tom~J Pollard, Lu~Shen, Li-wei~H Lehman, Mengling Feng, Mohammad Ghassemi, Benjamin Moody, Peter Szolovits, Leo Anthony~Celi, and Roger~G Mark. 2016.
\newblock Mimic-iii, a freely accessible critical care database.
\newblock \emph{Scientific data}, 3(1):1--9.

\bibitem[{Kposowa(2001)}]{kposowa2001unemployment}
Augustine~J Kposowa. 2001.
\newblock Unemployment and suicide: a cohort analysis of social factors predicting suicide in the us national longitudinal mortality study.
\newblock \emph{Psychological medicine}, 31(1):127--138.

\bibitem[{Kweon et~al.(2023)Kweon, Kim, Kim, Im, Cho, Bae, Oh, Lee, Moon, You et~al.}]{kweon2023publicly}
Sunjun Kweon, Junu Kim, Jiyoun Kim, Sujeong Im, Eunbyeol Cho, Seongsu Bae, Jungwoo Oh, Gyubok Lee, Jong~Hak Moon, Seng~Chan You, et~al. 2023.
\newblock Publicly shareable clinical large language model built on synthetic clinical notes.
\newblock \emph{arXiv preprint arXiv:2309.00237}.

\bibitem[{Lelkes et~al.(2023)Lelkes, Loreaux, Schuster, Chen, and Rajkomar}]{lelkes2023sdoh}
Adam~D Lelkes, Eric Loreaux, Tal Schuster, Ming-Jun Chen, and Alvin Rajkomar. 2023.
\newblock Sdoh-nli: a dataset for inferring social determinants of health from clinical notes.
\newblock \emph{arXiv preprint arXiv:2310.18431}.

\bibitem[{Lewis et~al.(2020)Lewis, Ott, Du, and Stoyanov}]{lewis-etal-2020-pretrained}
Patrick Lewis, Myle Ott, Jingfei Du, and Veselin Stoyanov. 2020.
\newblock \href {https://doi.org/10.18653/v1/2020.clinicalnlp-1.17} {Pretrained language models for biomedical and clinical tasks: Understanding and extending the state-of-the-art}.
\newblock In \emph{Proceedings of the 3rd Clinical Natural Language Processing Workshop}, pages 146--157, Online. Association for Computational Linguistics.

\bibitem[{Li et~al.(2023{\natexlab{a}})Li, Moon, Purkayastha, Celi, Trivedi, and Gichoya}]{li2023ethics}
Hanzhou Li, John~T Moon, Saptarshi Purkayastha, Leo~Anthony Celi, Hari Trivedi, and Judy~W Gichoya. 2023{\natexlab{a}}.
\newblock Ethics of large language models in medicine and medical research.
\newblock \emph{The Lancet Digital Health}, 5(6):e333--e335.

\bibitem[{Li et~al.(2023{\natexlab{b}})Li, Wang, and Yu}]{li2023two}
Rumeng Li, Xun Wang, and Hong Yu. 2023{\natexlab{b}}.
\newblock Two directions for clinical data generation with large language models: Data-to-label and label-to-data.
\newblock In \emph{Proceedings of the Conference on Empirical Methods in Natural Language Processing. Conference on Empirical Methods in Natural Language Processing}, volume 2023, page 7129. NIH Public Access.

\bibitem[{Liu et~al.(2019)Liu, Ott, Goyal, Du, Joshi, Chen, Levy, Lewis, Zettlemoyer, and Stoyanov}]{liu2019roberta}
Yinhan Liu, Myle Ott, Naman Goyal, Jingfei Du, Mandar Joshi, Danqi Chen, Omer Levy, Mike Lewis, Luke Zettlemoyer, and Veselin Stoyanov. 2019.
\newblock Roberta: A robustly optimized bert pretraining approach.
\newblock \emph{arXiv preprint arXiv:1907.11692}.

\bibitem[{Lybarger et~al.(2023{\natexlab{a}})Lybarger, Dobbins, Long, Singh, Wedgeworth, Uzuner, and Yetisgen}]{lybarger2023leveraging}
Kevin Lybarger, Nicholas~J Dobbins, Ritche Long, Angad Singh, Patrick Wedgeworth, {\"O}zlem Uzuner, and Meliha Yetisgen. 2023{\natexlab{a}}.
\newblock Leveraging natural language processing to augment structured social determinants of health data in the electronic health record.
\newblock \emph{Journal of the American Medical Informatics Association}, 30(8):1389--1397.

\bibitem[{Lybarger et~al.(2023{\natexlab{b}})Lybarger, Yetisgen, and Uzuner}]{lybarger20232022}
Kevin Lybarger, Meliha Yetisgen, and {\"O}zlem Uzuner. 2023{\natexlab{b}}.
\newblock The 2022 n2c2/uw shared task on extracting social determinants of health.
\newblock \emph{Journal of the American Medical Informatics Association}, 30(8):1367--1378.

\bibitem[{Majoka and Schimming(2021)}]{majoka2021effect}
Muniza~Anum Majoka and Corbett Schimming. 2021.
\newblock Effect of social determinants of health on cognition and risk of alzheimer disease and related dementias.
\newblock \emph{Clinical Therapeutics}, 43(6):922--929.

\bibitem[{Mitra et~al.(2025)Mitra, Chen, Liu, Kessler, and Yu}]{mitra2025post}
Avijit Mitra, Kun Chen, Weisong Liu, Ronald~C Kessler, and Hong Yu. 2025.
\newblock Post-discharge suicide prediction among us veterans using natural language processing-enriched social and behavioral determinants of health.
\newblock \emph{npj Mental Health Research}, 4(1):8.

\bibitem[{Mitra et~al.(2023)Mitra, Pradhan, Melamed, Chen, Hoaglin, Tucker, Reisman, Yang, Liu, Tsai et~al.}]{mitra2023associations}
Avijit Mitra, Richeek Pradhan, Rachel~D Melamed, Kun Chen, David~C Hoaglin, Katherine~L Tucker, Joel~I Reisman, Zhichao Yang, Weisong Liu, Jack Tsai, et~al. 2023.
\newblock Associations between natural language processing--enriched social determinants of health and suicide death among us veterans.
\newblock \emph{JAMA network open}, 6(3):e233079--e233079.

\bibitem[{Navigli et~al.(2023)Navigli, Conia, and Ross}]{navigli2023biases}
Roberto Navigli, Simone Conia, and Bj{\"o}rn Ross. 2023.
\newblock Biases in large language models: origins, inventory, and discussion.
\newblock \emph{ACM Journal of Data and Information Quality}, 15(2):1--21.

\bibitem[{Nelson et~al.(2020)Nelson, Bhutta, Harris, Danese, and Samara}]{nelson2020adversity}
Charles~A Nelson, Zulfiqar~A Bhutta, Nadine~Burke Harris, Andrea Danese, and Muthanna Samara. 2020.
\newblock Adversity in childhood is linked to mental and physical health throughout life.
\newblock \emph{bmj}, 371.

\bibitem[{Nicholson(2012)}]{nicholson2012review}
Nicholas~R Nicholson. 2012.
\newblock A review of social isolation: an important but underassessed condition in older adults.
\newblock \emph{The journal of primary prevention}, 33:137--152.

\bibitem[{Nijhawan et~al.(2019)Nijhawan, Metsch, Zhang, Feaster, Gooden, Jain, Walker, Huffaker, Mugavero, Jacobs et~al.}]{nijhawan2019clinical}
Ank~E Nijhawan, Lisa~R Metsch, Song Zhang, Daniel~J Feaster, Lauren Gooden, Mamta~K Jain, Robrina Walker, Shannon Huffaker, Michael~J Mugavero, Petra Jacobs, et~al. 2019.
\newblock Clinical and sociobehavioral prediction model of 30-day hospital readmissions among people with hiv and substance use disorder: beyond electronic health record data.
\newblock \emph{JAIDS Journal of Acquired Immune Deficiency Syndromes}, 80(3):330--341.

\bibitem[{Nori et~al.(2023)Nori, King, McKinney, Carignan, and Horvitz}]{nori2023capabilities}
Harsha Nori, Nicholas King, Scott~Mayer McKinney, Dean Carignan, and Eric Horvitz. 2023.
\newblock Capabilities of gpt-4 on medical challenge problems.
\newblock \emph{arXiv preprint arXiv:2303.13375}.

\bibitem[{Omar et~al.(2025)Omar, Sorin, Agbareia, Apakama, Soroush, Sakhuja, Freeman, Horowitz, Richardson, Nadkarni et~al.}]{omar2025evaluating}
Mahmud Omar, Vera Sorin, Reem Agbareia, Donald~U Apakama, Ali Soroush, Ankit Sakhuja, Robert Freeman, Carol~R Horowitz, Lynne~D Richardson, Girish~N Nadkarni, et~al. 2025.
\newblock Evaluating and addressing demographic disparities in medical large language models: a systematic review.
\newblock \emph{International Journal for Equity in Health}, 24(1):57.

\bibitem[{Omiye et~al.(2023)Omiye, Lester, Spichak, Rotemberg, and Daneshjou}]{omiye2023large}
Jesutofunmi~A Omiye, Jenna~C Lester, Simon Spichak, Veronica Rotemberg, and Roxana Daneshjou. 2023.
\newblock Large language models propagate race-based medicine.
\newblock \emph{NPJ Digital Medicine}, 6(1):195.

\bibitem[{Patra et~al.(2021)Patra, Sharma, Vekaria, Adekkanattu, Patterson, Glicksberg, Lepow, Ryu, Biernacka, Furmanchuk et~al.}]{patra2021extracting}
Braja~G Patra, Mohit~M Sharma, Veer Vekaria, Prakash Adekkanattu, Olga~V Patterson, Benjamin Glicksberg, Lauren~A Lepow, Euijung Ryu, Joanna~M Biernacka, Al’ona Furmanchuk, et~al. 2021.
\newblock Extracting social determinants of health from electronic health records using natural language processing: a systematic review.
\newblock \emph{Journal of the American Medical Informatics Association}, 28(12):2716--2727.

\bibitem[{Radford et~al.(2019)Radford, Wu, Child, Luan, Amodei, Sutskever et~al.}]{radford2019language}
Alec Radford, Jeffrey Wu, Rewon Child, David Luan, Dario Amodei, Ilya Sutskever, et~al. 2019.
\newblock Language models are unsupervised multitask learners.
\newblock \emph{OpenAI blog}, 1(8):9.

\bibitem[{Sahu et~al.(2022)Sahu, Rodriguez, Laradji, Atighehchian, Vazquez, and Bahdanau}]{sahu2022data}
Gaurav Sahu, Pau Rodriguez, Issam~H Laradji, Parmida Atighehchian, David Vazquez, and Dzmitry Bahdanau. 2022.
\newblock Data augmentation for intent classification with off-the-shelf large language models.
\newblock \emph{arXiv preprint arXiv:2204.01959}.

\bibitem[{Savova et~al.(2008)Savova, Ogren, Duffy, Buntrock, and Chute}]{savova2008mayo}
Guergana~K Savova, Philip~V Ogren, Patrick~H Duffy, James~D Buntrock, and Christopher~G Chute. 2008.
\newblock Mayo clinic nlp system for patient smoking status identification.
\newblock \emph{Journal of the American Medical Informatics Association}, 15(1):25--28.

\bibitem[{Schick and Sch{\"u}tze(2021)}]{schick-schutze-2021-generating}
Timo Schick and Hinrich Sch{\"u}tze. 2021.
\newblock \href {https://doi.org/10.18653/v1/2021.emnlp-main.555} {Generating datasets with pretrained language models}.
\newblock In \emph{Proceedings of the 2021 Conference on Empirical Methods in Natural Language Processing}, pages 6943--6951, Online and Punta Cana, Dominican Republic. Association for Computational Linguistics.

\bibitem[{Schmidgall et~al.(2024)Schmidgall, Harris, Essien, Olshvang, Rahman, Kim, Ziaei, Eshraghian, Abadir, and Chellappa}]{schmidgall2024addressing}
Samuel Schmidgall, Carl Harris, Ime Essien, Daniel Olshvang, Tawsifur Rahman, Ji~Woong Kim, Rojin Ziaei, Jason Eshraghian, Peter Abadir, and Rama Chellappa. 2024.
\newblock Addressing cognitive bias in medical language models.
\newblock \emph{arXiv preprint arXiv:2402.08113}.

\bibitem[{Service()}]{USDAERSF54:online}
USDA Economic~Research Service.
\newblock Food security in the u.s.
\newblock \url{https://www.ers.usda.gov/topics/food-nutrition-assistance/food-security-in-the-u-s/}.
\newblock (Accessed on 08/16/2024).

\bibitem[{Shonkoff(2012)}]{shonkoff2012committee}
Jack~P Shonkoff. 2012.
\newblock Committee on psychosocial aspects of child and family health; committee on early childhood adoption and dependent care; section on developmental and behavioral pediatrics (2012). the lifelong effects of early childhood adversity and toxic stress.
\newblock \emph{Pediatrics}, 129:e232.

\bibitem[{Singhal et~al.(2023{\natexlab{a}})Singhal, Azizi, Tu, Mahdavi, Wei, Chung, Scales, Tanwani, Cole-Lewis, Pfohl et~al.}]{singhal2023large}
Karan Singhal, Shekoofeh Azizi, Tao Tu, S~Sara Mahdavi, Jason Wei, Hyung~Won Chung, Nathan Scales, Ajay Tanwani, Heather Cole-Lewis, Stephen Pfohl, et~al. 2023{\natexlab{a}}.
\newblock Large language models encode clinical knowledge.
\newblock \emph{Nature}, 620(7972):172--180.

\bibitem[{Singhal et~al.(2023{\natexlab{b}})Singhal, Tu, Gottweis, Sayres, Wulczyn, Hou, Clark, Pfohl, Cole-Lewis, Neal et~al.}]{singhal2023towards}
Karan Singhal, Tao Tu, Juraj Gottweis, Rory Sayres, Ellery Wulczyn, Le~Hou, Kevin Clark, Stephen Pfohl, Heather Cole-Lewis, Darlene Neal, et~al. 2023{\natexlab{b}}.
\newblock Towards expert-level medical question answering with large language models.
\newblock \emph{arXiv preprint arXiv:2305.09617}.

\bibitem[{Takahashi et~al.(2015)Takahashi, Ryu, Olson, Winkler, Hathcock, Gupta, Sloan, Pathak, Bielinski, and Cerhan}]{takahashi2015health}
Paul~Y Takahashi, Euijung Ryu, Janet~E Olson, Erin~M Winkler, Matthew~A Hathcock, Ruchi Gupta, Jeff~A Sloan, Jyotishman Pathak, Suzette~J Bielinski, and James~R Cerhan. 2015.
\newblock Health behaviors and quality of life predictors for risk of hospitalization in an electronic health record-linked biobank.
\newblock \emph{International journal of general medicine}, pages 247--254.

\bibitem[{Thirunavukarasu et~al.(2023)Thirunavukarasu, Ting, Elangovan, Gutierrez, Tan, and Ting}]{thirunavukarasu2023large}
Arun~James Thirunavukarasu, Darren Shu~Jeng Ting, Kabilan Elangovan, Laura Gutierrez, Ting~Fang Tan, and Daniel Shu~Wei Ting. 2023.
\newblock Large language models in medicine.
\newblock \emph{Nature medicine}, 29(8):1930--1940.

\bibitem[{Tran et~al.(2024)Tran, Yang, Yao, and Yu}]{tran2024bioinstruct}
Hieu Tran, Zhichao Yang, Zonghai Yao, and Hong Yu. 2024.
\newblock Bioinstruct: instruction tuning of large language models for biomedical natural language processing.
\newblock \emph{Journal of the American Medical Informatics Association}, page ocae122.

\bibitem[{Truong et~al.(2020)Truong, Luke, Hammond, Wadhera, Reidhead, and Maddox}]{truong2020utilization}
Hannah~P Truong, Alina~A Luke, Gmerice Hammond, Rishi~K Wadhera, Mat Reidhead, and Karen E~Joynt Maddox. 2020.
\newblock Utilization of social determinants of health icd-10 z-codes among hospitalized patients in the united states, 2016--2017.
\newblock \emph{Medical care}, 58(12):1037--1043.

\bibitem[{Turner-Cobb et~al.(2000)Turner-Cobb, Sephton, Koopman, Blake-Mortimer, and Spiegel}]{turner2000social}
Julie~M Turner-Cobb, Sandra~E Sephton, Cheryl Koopman, Jane Blake-Mortimer, and David Spiegel. 2000.
\newblock Social support and salivary cortisol in women with metastatic breast cancer.
\newblock \emph{Psychosomatic Medicine}, 62(3):337--345.

\bibitem[{Uzuner et~al.(2008)Uzuner, Goldstein, Luo, and Kohane}]{uzuner2008identifying}
{\"O}zlem Uzuner, Ira Goldstein, Yuan Luo, and Isaac Kohane. 2008.
\newblock Identifying patient smoking status from medical discharge records.
\newblock \emph{Journal of the American Medical Informatics Association}, 15(1):14--24.

\bibitem[{Volkow and Blanco(2021)}]{volkow2021changing}
Nora~D Volkow and Carlos Blanco. 2021.
\newblock The changing opioid crisis: development, challenges and opportunities.
\newblock \emph{Molecular psychiatry}, 26(1):218--233.

\bibitem[{Wang et~al.(2024)Wang, Yao, Yang, Zhou, Li, Wang, Xu, and Yu}]{wang2024notechat}
Junda Wang, Zonghai Yao, Zhichao Yang, Huixue Zhou, Rumeng Li, Xun Wang, Yucheng Xu, and Hong Yu. 2024.
\newblock Notechat: a dataset of synthetic patient-physician conversations conditioned on clinical notes.
\newblock In \emph{Findings of the Association for Computational Linguistics ACL 2024}, pages 15183--15201.

\bibitem[{Wang et~al.(2023)Wang, Zhou, Zu, Xia, Chen, Zhang, Zheng, Ye, Zhang, Gui et~al.}]{wang2023instructuie}
Xiao Wang, Weikang Zhou, Can Zu, Han Xia, Tianze Chen, Yuansen Zhang, Rui Zheng, Junjie Ye, Qi~Zhang, Tao Gui, et~al. 2023.
\newblock Instructuie: Multi-task instruction tuning for unified information extraction.
\newblock \emph{arXiv preprint arXiv:2304.08085}.

\bibitem[{Wang et~al.(2015)Wang, Chen, Pakhomov, Arsoniadis, Carter, Lindemann, Sarkar, and Melton}]{wang2015automated}
Yan Wang, Elizabeth~S Chen, Serguei Pakhomov, Elliot Arsoniadis, Elizabeth~W Carter, Elizabeth Lindemann, Indra~Neil Sarkar, and Genevieve~B Melton. 2015.
\newblock Automated extraction of substance use information from clinical texts.
\newblock In \emph{AMIA Annual Symposium Proceedings}, volume 2015, page 2121. American Medical Informatics Association.

\bibitem[{Wang et~al.(2022)Wang, Kordi, Mishra, Liu, Smith, Khashabi, and Hajishirzi}]{wang2022self}
Yizhong Wang, Yeganeh Kordi, Swaroop Mishra, Alisa Liu, Noah~A Smith, Daniel Khashabi, and Hannaneh Hajishirzi. 2022.
\newblock Self-instruct: Aligning language models with self-generated instructions.
\newblock \emph{arXiv preprint arXiv:2212.10560}.

\bibitem[{Weng et~al.(2022)Weng, Zhu, Xia, Li, He, Liu, Sun, Liu, and Zhao}]{weng2022large}
Yixuan Weng, Minjun Zhu, Fei Xia, Bin Li, Shizhu He, Shengping Liu, Bin Sun, Kang Liu, and Jun Zhao. 2022.
\newblock Large language models are better reasoners with self-verification.
\newblock \emph{arXiv preprint arXiv:2212.09561}.

\bibitem[{Xu et~al.(2023)Xu, Cui, Yu, Kan, Shi, Zhuang, Jin, Ho, and Yang}]{xu2023knowledge}
Ran Xu, Hejie Cui, Yue Yu, Xuan Kan, Wenqi Shi, Yuchen Zhuang, Wei Jin, Joyce Ho, and Carl Yang. 2023.
\newblock Knowledge-infused prompting improves clinical text generation with large language models.
\newblock In \emph{NeurIPS 2023 Workshop on Synthetic Data Generation with Generative AI}.

\bibitem[{Yang et~al.(2022)Yang, Soltan, and Clifton}]{yang2022machine}
Jenny Yang, Andrew~AS Soltan, and David~A Clifton. 2022.
\newblock Machine learning generalizability across healthcare settings: insights from multi-site covid-19 screening.
\newblock \emph{NPJ digital medicine}, 5(1):69.

\bibitem[{Yang et~al.(2024)Yang, Mitra, Kwon, and Yu}]{yang2024clinicalmamba}
Zhichao Yang, Avijit Mitra, Sunjae Kwon, and Hong Yu. 2024.
\newblock Clinicalmamba: A generative clinical language model on longitudinal clinical notes.
\newblock \emph{arXiv preprint arXiv:2403.05795}.

\bibitem[{Yoo et~al.(2021)Yoo, Park, Kang, Lee, and Park}]{yoo2021gpt3mix}
Kang~Min Yoo, Dongju Park, Jaewook Kang, Sang-Woo Lee, and Woomyeong Park. 2021.
\newblock Gpt3mix: Leveraging large-scale language models for text augmentation.
\newblock \emph{arXiv preprint arXiv:2104.08826}.

\bibitem[{Yoon et~al.(2023)Yoon, Mizrahi, Ghalaty, Jarvinen, Ravi, Brune, Kong, Anderson, Lee, Meir et~al.}]{yoon2023ehr}
Jinsung Yoon, Michel Mizrahi, Nahid~Farhady Ghalaty, Thomas Jarvinen, Ashwin~S Ravi, Peter Brune, Fanyu Kong, Dave Anderson, George Lee, Arie Meir, et~al. 2023.
\newblock Ehr-safe: generating high-fidelity and privacy-preserving synthetic electronic health records.
\newblock \emph{NPJ Digital Medicine}, 6(1):141.

\bibitem[{Yu et~al.(2022)Yu, Yang, Guo, Bian, and Wu}]{yu2022assessing}
Zehao Yu, Xi~Yang, Yi~Guo, Jiang Bian, and Yonghui Wu. 2022.
\newblock Assessing the documentation of social determinants of health for lung cancer patients in clinical narratives.
\newblock \emph{Frontiers in public health}, 10:778463.

\bibitem[{Zack et~al.(2024)Zack, Lehman, Suzgun, Rodriguez, Celi, Gichoya, Jurafsky, Szolovits, Bates, Abdulnour et~al.}]{zack2024assessing}
Travis Zack, Eric Lehman, Mirac Suzgun, Jorge~A Rodriguez, Leo~Anthony Celi, Judy Gichoya, Dan Jurafsky, Peter Szolovits, David~W Bates, Raja-Elie~E Abdulnour, et~al. 2024.
\newblock Assessing the potential of gpt-4 to perpetuate racial and gender biases in health care: a model evaluation study.
\newblock \emph{The Lancet Digital Health}, 6(1):e12--e22.

\bibitem[{Zaratiana et~al.(2023)Zaratiana, Tomeh, Holat, and Charnois}]{zaratiana2023gliner}
Urchade Zaratiana, Nadi Tomeh, Pierre Holat, and Thierry Charnois. 2023.
\newblock Gliner: Generalist model for named entity recognition using bidirectional transformer.
\newblock \emph{arXiv preprint arXiv:2311.08526}.

\end{thebibliography}

\clearpage
\appendix

\section{Synth-SBDH}
\label{app_synth}

\subsection{SBDH Categories in Synth-SBDH}
\label{app_ctg}
The 14 SBDH categories and `pain' with their descriptions and sample text spans are shown in Table \ref{synth_sbdh_ctg}.
\begin{table*}
\centering
\fontsize{8pt}{8pt}\selectfont
\begin{threeparttable}
\begin{tabular}{lll}
\toprule
\textbf{SBDH categories} & \textbf{Definition}   & \textbf{Example text spans}                                                           \\
\midrule
\midrule
\textbf{Social Determinants} \\
\midrule
Social isolation      & \makecell[cl]{A state in which the individual lacks a sense of \\belonging socially, lacks engagement with others, \\has a minimal number of social contacts, and they \\are deficient in fulfilling and quality relationships.}    & Alone, lonely, etc.                                       \\\midrule
Physical isolation      & \makecell[cl]{Physical isolation results in less involvement with \\others, often due to disability, illness, housebound or \\bedbound, that prevents active participation in life \\outside of the home/immediate physical environment.
}    & Bedridden, housebound, etc. \\\midrule
Barriers to care      & \makecell[cl]{Barriers to care are factors that interfere with health-\\care access, and may include transportation issues, \\cognitive or communication difficulties, lack of trust \\in the care system, or lack of rapport with provider(s).} & \makecell[cl]{Transportation issues, \\communication problems etc.}     \\\midrule
Financial insecurity  & \makecell[cl]{The anxiety produced by the possible exposure to \\adverse economic events and the anticipation of the \\difficulty of recovering from them.}  & \makecell[cl]{Poor, low income, etc. }\\\midrule
Job   insecurity  & \makecell[cl]{Job insecurity includes unemployment, underemploy-\\ment, unstable employment, fear of losing a job or \\benefits, and vocational rehabilitation/training.}  & \makecell[cl]{Unemployed, lost job etc.} \\\midrule
Loss of relationship & \makecell[cl]{A loss of a personal relationship, including divorce, \\separation, death, estrangement, or breakdown of \\interpersonal communication.} & Divorce, widow etc. \\\midrule
Housing insecurity     & \makecell[cl]{Housing insecurity refers to unstable housing due to \\a variety of reasons which may include eviction, in-\\ability to afford rent, foreclosure, or displacement \\due to domestic/roommate/landlord issues.
} & Eviction, homeless, etc.\\\midrule
Food insecurity       & \makecell[cl]{Lack of consistent access to enough food for every \\person in a household to live an active, healthy life.}  & \makecell[cl]{Hungry, pantry, starvation, \\food voucher etc.} \\\midrule
Violence                & \makecell[cl]{The violence category includes elements of the \\individual's environment, as well as the larger societal \\environment. The presence of weapons, various types \\of abuse (physical, emotional/psychological, sexual), \\exposure to combat, bullying, harassment, threats, and \\racism\tnote{$\dagger$} are categorized as violence. Violence includes \\cases of both perpetrators and victims.} & \makecell[cl]{Firearms, violence, assault, \\abuse,   homicidal, racism etc.} \\\midrule
Legal problems        & \makecell[cl]{Legal problems entail violations of law, associated pun-\\ishments, and mention of related officials, places, and \\processes e.g., attorney, judge, parole officer, court, jail, \\prison, incarceration, child custody/child support issues.}  & \makecell[cl]{Imprisonment, parole, arrested, \\felony, prison etc.} \\
\midrule
\textbf{Behavioral Determinants} \\
\midrule
Substance abuse       & \makecell[cl]{Substance Abuse (marijuana excluded) covers the use \\of both legal (alcohol, tobacco) and illicit substances, \\addiction, substance abuse treatment/rehab/sobriety \\groups, and relapse.}  & \makecell[cl]{Alcohol, tobacco, cocaine, \\heroin, smoking, overdose etc.}\\\midrule
\makecell[cl]{Psychiatric   symptoms \\or disorders} & \makecell[cl]{Psychiatric symptoms or disorders category includes \\emotional/psychological difficulties and conditions that \\affect the ability to function well in daily life.} & \makecell[cl]{PTSD, depression, insomnia, \\schizophrenia, hallucination etc.} \\\midrule
Patient disability    & \makecell[cl]{The patient disability category includes impairments that \\affect daily life as evidenced by the presence of assistive \\devices, disability payments, and military service-\\connected ratings.
}  & Disabled, blind, wheelchair   etc.\\\midrule
\textbf{Others}\\\midrule
Pain
& \makecell[cl]{The pain category considers acute and chronic pain, \\arthralgia, migraine, and evidence of pain through \\mention of pain management/mitigation.}      & Pain, suffering, hurting, etc.\\ \midrule 
\\Transition of care    & \makecell[cl]{The transitions of care category identifies healthcare-\\related points of vulnerability; examples include \\admission, discharge, medication change, and change \\of provider.} & \makecell[cl]{Discharge, admission, change \\in medication,   transfer etc.}              \\
\bottomrule
\end{tabular}
\end{threeparttable}
\caption{List of all SBDH Categories}
\label{synth_sbdh_ctg}
\end{table*}

\subsection{Prompt}
\label{app_prompt}
We used the following prompt to generate 20 examples at a time from GPT-4. 
\begin{Verbatim}[breaklines=true]
The social determinants of health (SDOH) are the non-medical factors that influence health outcomes. They are the conditions in which people are born, grow, work, live, and age, and the wider set of forces and systems shaping the conditions of daily life. SDOHs have a major impact on people’s health, well-being, and quality of life. SDOHs encompass factors such as socioeconomic status, access to healthy food, education, housing, and physical environment, to name a few. Together with behavioral factors such as substance abuse, we get Social and behavioral determinants of health (SBDH). Below are 15 SBDH categories with definitions that we will consider.

1. Food Insecurity: Lack of consistent access to enough food for every person in a household to live an active, healthy life.
2. Job Insecurity: Job insecurity includes unemployment, underemployment, unstable employment, fear of losing a job or benefits, and vocational rehabilitation/training.
3. Housing Insecurity: Housing insecurity refers to unstable housing due to a variety of reasons which may include eviction, inability to afford rent, foreclosure, or displacement due to domestic/roommate/landlord issues.

... 

15 Psychiatric Symptoms or Disorders: Psychiatric Symptoms or Disorders category includes emotional/psychological difficulties and conditions that affect the ability to function well in daily life.

Below are a few sample examples with SBDH collected from different clinical notes where each example has six sections - `Text', `Textspan', `Reasoning', `SBDH', `Presence', and `Period'. `Text' contains the example with mention(s) of SBDH(s). `Textspan' consists of text spans from the `Text' with indications of SBDH separated by double vertical lines (||). `Reasoning' has the rationales or reasonings behind the SBDH annotations and follows the same order as in `Textspan'. Next, we have the SBDH category, its presence - yes or no, and period - current (exists currently) or history (events from the past).

Example 1.
Text: Patient's Seroquel dose remains the same.
Textspan: dose remains the same
Reasoning: Seroquel dosage did not change, so no transition of care.
SBDH: Transitions of Care
Presence: no
Period: current

Example 2.
Text: Patient has tobacco use disorder (in past medical history list)
Textspan: tobacco use disorder
Reasoning: Tobacco use is categorized as substance abuse; patient has history of tobacco use.
SBDH: Substance Abuse
Presence: yes
Period: history

Example 3.
Text: Veteran is anticipated to discharge home once medically stable for discharge transported by his wife.
Textspan: discharge || home || transported by his wife
Reasoning: Discharge from the hospital is considered as transition of care. || Veteran has a home to go after discharge, so he is less likely to have housing problems. || Having family or friends to transport the patient is absence of barriers to care.
SBDH: Transitions of Care || Housing Insecurity || Barriers to Care
Presence: yes || no || no
Period: current || current || current

... 

Example 10.
Text: Consult for polysubstance abuse, in particular OUD with fentanyl use. 
Textspan: polysubstance abuse || OUD || fentanyl use
Reasoning: Polysubstance abuse outside their intended use is considered substance abuse. || Opioid use disorder (OUD) is considered substance abuse. || Using illicit or prescription drugs (fentanyl) outside their intended use is considered substance abuse.
SBDH: Substance Abuse || Substance Abuse || Substance Abuse
Presence: yes || yes || yes
Period: current || current || current

Your task is to generate examples with mentions of SBDH, provide your reasonings behind them, and finally annotate those mentions in the examples. The requirements are:
1. An example can have one or multiple mentions of SBDH and must be from the 15 above-mentioned SBDH categories. 
2. More importantly, each example should emulate a text snippet from a patient's electronic health records with no more than three sentences. 
3. You must make the examples (specially 'Text' and 'Reasoning' sections) as diverse as possible, both syntactically and semantically. Do not start examples with the same pattern.

Now, generate 20 examples that meet the requirements mentioned above. Format examples as a valid JSON with the following structure:
[
    {
        `Text':..., 
        `Annotations': [
            {
                `Textspan':...,
                `Reasoning':...,
                `SBDH':...,
                `Presence':...,
                `Period':...,
            },
            {
                `Textspan':...,
                `Reasoning':...,
                `SBDH':...,
                `Presence':...,
                `Period':...,
            },
            ...
        ]
    },
    {
        `Text':...,
        `Annotations': [
            {
                `Textspan':...,
                `Reasoning':...,
                `SBDH':...,
                `Presence':...,
                `Period':...,
            },
            {
                `Textspan':...,
                `Reasoning':...,
                `SBDH':...,
                `Presence':...,
                `Period':...,
            },
            ...
        ]
    },
    ...
]
\end{Verbatim}
\subsection{Seed examples}
\label{app_seed}
We used 45 expert-written seed examples to mimic the style and tone in EHR notes. These were utilized to prompt GPT-4 to ensure that the generated examples maintain a similar language. Every time we prompt GPT-4, we randomly sample 10 examples and use them in the prompt as shown in section \ref{app_prompt}. The list of 45 examples is shown below. Here multiple annotations and rationales (reasoning) from the same text are separated by `||'. Each annotation in `Annotations' is in (Text span, SBDH label, Presence label, Period label) format. 
\begin{Verbatim}[breaklines=true]
Example 1.
Text: He explained that he was hungry and did not have a lot of money so took about $60 worth of food.
Reasoning: He was hungry and did not have enough money to buy food. || He did not have enough money to buy food. || He stole food suggesting he did not have any to eat.
Annotations: (hungry, Food Insecurity, yes, current) || (not have a lot of money, Financial Insecurity, yes, current) || ($60 worth of food, Food Insecurity, yes, current)

Example 2.
Text: Denies substance abuse and legal issues, and vet states he has plenty of food to eat.
Reasoning: Veteran is denying subtance abuse. || Vetaran is denying any legal problems. || Vetaran states that he does not have any shortage of food.
Annotations: (substance abuse, Substance Abuse, no, current) || (legal issues, Legal Problems, no, current) || (plenty of food to eat, Food Insecurity, no, current)

Example 3.
Text: Patient reports she was assaulted by her domestic partner today and needs help finding a place to stay.
Reasoning: Patient seeking help after being assaulted by domestic partner. Assault is categorized as violence. || Assault has disrupted relationship with domestic partner. || Patient needs a different place to live after being assaulted by domestic partner. 
Annotations: (assaulted, Violence, yes, current) || (by her domestic partner, Loss of Relationship, yes, current) || (needs help finding a place to stay, Housing Insecurity, yes, current)

Example 4.
Text: He was provided with a food box and two bus tickets.
Reasoning: Receiving a food box indicates that patient is currently experiencing food insecurity. || Receiving bus tickets suggests that patient is currently having an issue with transportation, which is categorized as barrier to care.
Annotations: (provided with a food box, Food Insecurity, yes, current) || (two bus tickets, Barriers to Care, yes, current)

Example 5.
Text: She is keeping herself busy attending church and has met some individuals who are supportive.
Reasoning: She attends church, where she is with others. || She has met people who are supportive, and is currently not experiencing social isolation.
Annotations: (attending church, Social Isolation, no, current) || (individuals who are supportive, Social Isolation, no, current)

Example 6.
Text: These parents also started giving her crystal meth at age 12 and forced her out of the home at age 14. 
Reasoning: Patient's parents gave her crystal meth as a child, which disrupted the parent/child relationship. || Crystal meth is an illicit substance; its use is categorized as substance abuse || Forcing patient out of the home as a child resulted in housing insecurity.
Annotations: (parents, Loss of Relationship, yes, history) || (giving her crystal meth, Substance Abuse, yes, history) || (forced her out of the home, Housing Insecurity, yes, history)

Example 7.
Text: Veteran has little social support and very little contact with family. He is now completely bedridden, weaker, on puree diet, incontinent.
Reasoning: Veteran doesn't have much social support. || Veteran does not have much contact with family. || Veteran is confined to bed and weak.
Annotations: (little social support, Social Isolation, yes, current) || (very little contact with family, Social Isolation, yes, current) || (completely bedridden, Physical Isolation, yes, current)

Example 8.
Text: Veteran is anticipated to discharge home once medically stable for discharge transported by his wife.
Reasoning: Discharge from the hospital is considered as transition of care. || Veteran has a home to go after discharge, so he is less likely to have housing problems. || Having family or friends to transport the patient is absence of barriers to care.
Annotations: (discharge, Transitions of Care, yes, current) || (home, Housing Insecurity, no, current) || (transported by his wife, Barriers to Care, no, current)

Example 9.
Text: Veteran is currently wheelchair bound, and has not ambulated in over a year due to multiple medical and physical issues. Assisted living facility reports that he has not moved from bed due to severe right hip pain.
Reasoning: Patient is wheelchair bound and hasn't walked in over a year due to his disability. || Patient has severe hip pain.
Annotations: (wheelchair bound, Patient Disability, yes, current) || (severe right hip pain, Pain, yes, current)

Example 10.
Text: In terms of paranoia, patient appears to have some residual paranoia both about people outside and inside the hospital but does not actively wish to harm anyone or have homicidal ideation towards any one person.
Reasoning: Patient has some residual paranoia, a sign of psychiatric problems. ||Patient does not intend to harm anyone - an absence of violent behavior. || Specific mention of the absence of homicidal thoughts.
Annotations: (paranoia, Psychiatric Symptoms or Disorders, yes, current) || (does not actively wish to harm anyone, Violence, no, current) || (homicidal ideation, Violence, no, current)

Example 11.
Text: Patient says she was incarcerated for parole violation on an armed robbery charge.
Reasoning: Patient was incarcerated for a violation of the law. || Parole violation is a violation of law and may cause legal issues. || Armed robbery is an act of criminal activity, causing a violation of law.
Annotations: (incarcerated, Legal Problems, yes, history) || (parole violation, Legal Problems, yes, history) || (armed robbery charge, Legal Problems, yes, history)

Example 12.
Text: Is the patient "marginally housed"? No. Has patient been homeless within the past year for any length of time? No.
Reasoning: Patient is not currently "marginally housed". || Patient has not been homeless for any length of time during the past year.
Annotations: (marginally housed, Housing Insecurity, no, current) || (homeless, Housing Insecurity, no, current)

Example 13.
Text: Major Depression, recurrent, severe; originally presented with SI to walk into traffic; no SI currently but still not able to contact for safety outside of hospital and wants to be transferred to Psychiatry unit.
Reasoning: Major Depression is a psychiatric symptom. || Hospitalization is categorized as transition of care; patient is hospitalized. || Being transferred to another department/unit is categorized as transition of care; patient is requesting to be transferred. || Being transfered to psychiatry unit means inpatient psychiatry admission, which is possibly due to any psychiatric symptoms or disorders.
Annotations: (Major Depression, Psychiatric Symptoms or Disorders, yes, current) || (outside of hospital, Transitions of Care, yes, current) || (wants to be transferred, Transitions of Care, yes, current) || (to Psychiatry unit, Psychiatric Symptoms or Disorders, yes, current) 

Example 14.
Text: The veteran's daughter reports that the veteran does not use alcohol, tobacco, or illicit substances.
Reasoning: Veteran does not consume alcohol. || Veteran does not use tobacco. || Veteran does not use illicit substances.
Annotations: (alcohol, Substance Abuse, no, current) || (tobacco, Substance Abuse, no, current) || (illicit substances, Substance Abuse, no, current)

Example 15.
Text: He says he has lost his job again and feels ashamed to even speak on the phone to his daughters.
Reasoning: Losing job is a clear sign of job insecurity.
Annotations: (lost his job, Job Insecurity, yes, current)

Example 16.
Text: Veteran spoke of her very satisfying marriage, and her love of her husband.
Reasoning: Veteran has a satisfying marriage. || Veteran loves her husband.
Annotations: (very satisfying marriage, Loss of Relationship, no, current) || (love of her husband, Loss of Relationship, no, current)

Example 17.
Text: Unintentional weight loss - stress and food insecurity related.
Reasoning: Stress is a psychological symptom. || Weight loss due to food insecurity was mentioned.
Annotations: (stress, Psychiatric Symptoms or Disorders, yes, current) || (food insecurity, Food Insecurity, yes, current)

Example 18.
Text: Vocational rehab will assist Veteran in job search activity bringing job leads, applications, and computer assistance.
Reasoning: Wording suggests that veteran needs assistance with finding employment. || Wording indicates veteran needs assistance with finding employment.
Annotations: (job search, Job Insecurity, yes, current) || (job leads, Job Insecurity, yes, current)

Example 19.
Text: Patient has worked most of his adult life as a cook and is employed at a restaurant as a chef.
Reasoning: Patient was employed most of his adult life, so little to no chance of past employment insecurity. || Patient is currently employed, suggesting no employment insecurity.
Annotations: (worked, Job Insecurity, no, history) || (employed, Job Insecurity, no, current)

Example 20.
Text: VASH social worker meets with veteran to ensure veteran continues developing and maintaining treatment plan goals of maintaining housing.
Reasoning: VASH is VA subsidized housing - any patient receiving VASH has housing insecurity. || "maintain housing" is a phrasing indicative of housing insecurity because it shows it is a current concern.
Annotations: (VASH, Housing Insecurity, yes, current) || (maintaining housing, Housing Insecurity, yes, current)

Example 21.
Text: Living situation - rental - lives with father
Reasoning: Patient is at a rental property with no indication of any housing issues.|| Lives with father, so no housing insecurity.
Annotations: (rental, Housing Insecurity, no, current) || (lives with father, Housing Insecurity, no, current)

Example 22.
Text: Veteran endorsed recent financial stressor which includes an old debt that was never paid.
Reasoning: Patient mentions financial stressors, suggesting financial insecurity. || Mentions of debt is can be highly correlated with financial insecurity.
Annotations: (financial stressor, Financial Insecurity, yes, current) || (debt that was never paid, Financial Insecurity, yes, current)

Example 23.
Text: Bouts of depression off and on, currently not wanting to be around others, isolating and reports feeling depressed every day of the last two weeks at least.
Reasoning: Depression is a psychiatric symptom. || Self-isolation is an indicator of social isolation. || Not wanting to be around others is social isolation. || The patient reports feeling depressed and depression is a psychiatric symptom.
Annotations: (depression, Psychiatric Symptoms or Disorders, yes, current) || (not wanting to be around others, Social Isolation, yes, current) || (isolating, Social Isolation, yes, current) || (depressed, Psychiatric Symptoms or Disorders, yes, current)

Example 24.
Text: Reports stressors including housing insecurity, chronic feelings of hopelessness, and poor social support.
Reasoning: Housing insecurity is mentioned as a stressor. || Feeling hopeless is a psychiatric symptom. || Mentions of poor or nonexistent social support are social isolation.
Annotations: (housing insecurity, Housing Insecurity, yes, current) || (hopelessness, Psychiatric Symptoms or Disorders, yes, current) || (social support, Social Isolation, yes, current))

Example 25.
Text: Veteran relates that she feels like a lot of her stress is related to marital dynamic and asks for marital counseling.
Reasoning: Stress due to marital dynamic can be indicative of a breakdown in communication or support. || Any mention of relationship counseling is a loss of relationship as it suggests a breakdown in communication or support.
Annotations: (marital dynamic, Loss of Relationship, yes, current) || (marital counseling, Loss of Relationship, yes, current)

Example 26.
Text: Veteran is reporting unwanted thoughts of harming two brothers as evidenced by him stating "they jumped me".
Reasoning: Thoughts of harming others is considered violence. || Being jumped is a physical component of violence.
Annotations: (thoughts of harming, Violence, yes, current) || (jumped me, Violence, yes, current)

Example 27.
Text: Patient had four inpatient psychiatric hospitalizations from several VA hospitals, and his most recent admission was for SI.
Reasoning: Past psychiatric hospitalization must be due to psychiatric concerns. || Any mention of previous hospitalization is considered a transition of care. || Any mention of admission is considered a transition of care. This happened in the past.
Annotations: (psychiatric hospitalization, Psychiatric Symptoms or Disorders, yes, history) || (hospitalization, Transitions of Care, yes, history) || (admission, Transitions of Care, yes, history)

Example 28.
Text: Admitting diagnoses/problems: abdominal pain. 
Reasoning: Patient has abdominal pain.
Annotations: (pain, Pain, yes, current)

Example 29.
Text: Veteran no longer requires a walker for ambulating.
Reasoning: Not requiring assistive devices shows that patient no longer has a disability.
Annotations: (walker, Patient Disability, no, current) 

Example 30.
Text: Consult for polysubstance abuse, in particular OUD with fentanyl use.
Reasoning: Polysubstance abuse outside their intended use is considered substance abuse. || Opioid use disorder (OUD) is considered substance abuse. || Using illicit or prescription drugs (fentanyl) outside their intended use is considered substance abuse.
Annotations: (polysubstance abuse, Substance Abuse, yes, current) || (OUD, Substance Abuse, yes, current) || (fentanyl use, Substance Abuse, yes, current) 

Example 31.
Text: Unemployed but reported is seeking work.
Reasoning: Patient is currently unemployed || Patient is looking for a job.
Annotations: (unemployed, Job Insecurity, yes, current) || (seeking work, Job Insecurity, yes, current)

Example 32.
Text: Veteran was evicted from his most recent apartment for non-payment of rent. He will be homeless at discharge.
Reasoning: Eviction is categorized as housing insecurity. || Loss of recent apartment means that patient currently has no home. || Non-payment of rent is categorized as housing insecurity. || Patient will be homeless when discharged, meaning housing insecurity. || Discharge is categorized as transition of care.
Annotations: (was evicted, Housing Insecurity, yes, current) || (most recent apartment, Housing Insecurity, yes, current) || (non-payment of rent, Housing Insecurity, yes, current) || (will be homeless, Housing Insecurity, yes, current) || (at discharge, Transitions of Care, yes, current)

Example 33.
Text: He is very worried about finances, including his children's back to school expenses and fees.
Reasoning: He is currently worried about finances which. || His current financial concerns include his children's school expenses. || He is currently worried about his children's school fees due to finanacial issues.
Annotations: (worried about finances, Financial Insecurity, yes, current) || (school expenses, Financial Insecurity, yes, current) || (fees, Financial Insecurity, yes, current)

Example 34.
Text: This writer received a call from the veteran inquiring about the VJO program as he has a misdemeanor case for theft.
Reasoning: VJO is acronym for Veterans Justice Outreach which indicates veteran has a legal issue. || Misdemeanor is a less serious criminal offense || Theft is a criminal offense.
Annotations: (VJO, Legal Problems, yes, current) || misdemeanor case, Legal Problems, yes, current) || (theft, Legal Problems, yes, current)

Example 35.
Text: Per son, veteran's prognosis remains poor. He is completely bedbound and requires help with all ADLs.
Reasoning: Veteran is confined to bed, and unable to complete activities of daily living.
Annotations: (completely bedbound, Physical Isolation, yes, current)

Example 36.
Text: Veteran reports that she actively communicates with mental health provider and has been complying with treatment recommendations and medications.
Reasoning: Veteran communicates with provider and complies with treatment. There is no barrier to care.
Annotations: (actively communicates, Barriers to Care, no, current)

Example 37.
Text: Veteran has recently begun to describe his exposure to military war experiences, i.e. burning bodies, fellow soldiers killed. 
Reasoning: Veteran has recently been describing violence he was exposed to during military combat. || Veteran saw burning bodies when serving in the military. || Veteran saw fellow soldiers killed while serving in the military.
Annotations: (military war experiences, Violence, yes, history) || (burning bodies, Violence, yes, history) || (fellow soldiers killed, Violence, yes, history)

Example 38.
Text: Patient's Seroquel dose remains the same.
Reasoning: Seroquel dosage did not change, so no transition of care.
Annotations: (dose remains the same, Transitions of Care, no, current)

Example 39.
Text: Patient's pain has been addressed by patient's provider and pain interventions have already been ordered.
Reasoning: Patient is experiencing pain. || Patient's pain is being addressed with intervention.
Annotations: (pain, Pain, yes, current) || (pain, Pain, yes, current)

Example 40.
Text: Veteran needs assistance with setting up since he is blind.
Reasoning: Patient is blind. Blindness is a disability.
Annotations: (he is blind, Patient Disability, yes, current)

Example 41.
Text: Veteran struggles with sleep apnea and PTSD which causes sleep issues.
Reasoning: Veteran suffers from PTSD, which is categorized as a psychiatric symptom. || Veteran has sleep issues; sleep issues are categorized as psychiatric symptom.
Annotations: (PTSD, Psychiatric Symptoms or Disorders, yes, current) || (causes sleep issues, Psychiatric Symptoms or Disorders, yes, current)

Example 42.
Text: Patient has tobacco use disorder (in past medical history list)
Reasoning: Tobacco use is categorized as substance abuse; patient has history of tobacco use.
Annotations: (tobacco use disorder, Substance Abuse, yes, history)

Example 43.
Text: The client self-reports that his rent and utilities are paid and up to date.
Reasoning: Mentions of rent being paid is considered an absence of financial insecurity. || Mentions of any bills being paid is considered an absence of financial insecurity.
Annotations: (rent, Financial Insecurity, no, current) || (utilities are paid, Financial Insecurity, no, current)

Example 44.
Text: Reports losing the trailer after losing job and getting evicted.
Reasoning: Losing a property (trailer) can be attributed to financial insecurity. || Losing a job is a sign of job insecurity. || Eviction is categorized as housing insecurity.
Annotations: (losing the trailer, Financial Insecurity, yes, current) || (losing job, Job Insecurity, yes, current) || (evicted, Housing Insecurity, yes, current)

Example 45.
Text: Social worker will continue to follow veteran for any further social work or discharge planning needs during this admission.
Reasoning: Any mention of discharge is considered a transition of care. || Any mention of admission is considered a transition of care.
Annotations: (discharge, Transitions of Care, yes, current) || (admission, Transitions of Care, yes, current)
\end{Verbatim}

\section{Datasets}
\label{app_data}
\subsection{MIMIC-SBDH\texorpdfstring{$_\text{aligned}$}{_aligned}}
MIMIC-SBDH \citep{ahsan2021mimic} provides SBDH annotations for the social history sections of 7,025 discharge summaries from MIMIC-III. It contains 7 SBDH categories (binary or categorical) - `Community', `Education', `Economics', `Environment', `Alcohol Use', `Tobacco Use', and `Drug Use'. Definitions of these SBDH categories and their value sets are provided in the original dataset paper. 

In MIMIC-SBDH, SBDH detection is formulated as a category-specific classification task, wherein separate classifiers are trained for each SBDH category. However, in practical applications, it is more desirable to train a single model capable of detecting multiple categories simultaneously (multi-label classification), as this approach is more computationally and time efficient. To facilitate this, we aligned the MIMIC-SBDH annotations with Synth-SBDH and derived four binary classification categories based on the original annotations and definitions: -
\begin{enumerate}
    \item `Substance abuse': This category aggregates three SBDH categories — \textit{Alcohol Use}, \textit{Tobacco Use}, and \textit{Drug Use} — representing illicit or harmful substance use. A value of \code{Present} in any of these categories is mapped to 1; all other values are mapped to 0.
    \item `Housing insecurity': In the \textit{Environment} category, a value of False indicates that the patient was homeless. We rename this category as `Housing Insecurity' and assign a value of 1 to \code{False}, and 0 otherwise.
    \item `Financial insecurity': The \textit{Economics} category is treated as indicative of employment status, where False implies the patient was unemployed. We rename this category to `Financial Insecurity' and map \code{False} to 1, and all other values to 0.
    \item `Isolation or loss of relationship': The \textit{Community} category includes two subcategories - \textit{Community-present} and \textit{Community-absent} - reflecting the presence or absence of social support, respectively. We focus on \textit{Community-absent} and map \code{True} to 1, and \code{False} to 0.
\end{enumerate}
MIMIC-SBDH with these modified categories constitutes MIMIC-SBDH$_\text{aligned}$. We chose a 70:10:20 split to create training, development, and test sets; yielding 4,917, 702, and 1,406 examples respectively. The distributions of SBDH categories across all splits are shown in Table \ref{tbl_mimic_sbdh_ctg}. Note that there are 1,285, 173, and 309 examples with no mention of SBDH in training, development, and test sets respectively.

\begin{table}
\fontsize{8pt}{8pt}\selectfont
\centering
\begin{threeparttable}
\begin{tabular}{lrrr}
\toprule
SBDH categories & Training &Development &Test \\
\midrule\midrule
Substance abuse       & 1,848 & 274 & 547  \\\midrule
Housing insecurity   & 41 & 4 & 18 \\\midrule
Financial insecurity  & 1,190 & 174 & 378 \\\midrule
\makecell[l]{Isolation or \\loss of relationship}  & 553 & 77 & 154 \\\midrule\midrule
Total & 3,632 & 529 & 1,097 \\
\bottomrule
\end{tabular}
\end{threeparttable}
\caption{SBDH category distribution for MIMIC-SBDH$_\text{aligned}$ for the MLC task.}
\label{tbl_mimic_sbdh_ctg}
\end{table}

\subsection{VA-SBDH}
\label{app_va_sbdh}
VA-SBDH \citep{mitra2023associations} has span-level annotations for 12 SBDH categories - `Isolation or loss of relationship', `Transition of care', `Barriers to care', `Financial or job insecurity', `Housing instability', `Food insecurity', `Violence', `Legal problems', `Substance abuse', `Psychiatric symptoms or disorders', and `Patient disability'. All categories follow similar definitions as listed in Table \ref{synth_sbdh_ctg} except the following two categories - 
\begin{enumerate}
    \item `Isolation or loss of relationship' - This category combines `Social isolation', `Physical isolation' and `Loss of relationship'.
    \item `Financial or job insecurity' - This category combines `Financial insecurity' and `Job insecurity'.
\end{enumerate} 
VA-SBDH comes with pre-defined training, development, and test splits with 12,236, 4,163, and 4,171 examples respectively. Because of span-level annotation, VA-SBDH can be utilized for both MLC and NER tasks. The annotation distributions of 12 SBDH categories across all splits are shown in Table \ref{tbl_va_sbdh_ctg}.

\begin{table}
\fontsize{8pt}{8pt}\selectfont
\centering
\begin{threeparttable}
\begin{tabular}{lrrr}
\toprule
SBDH categories & Training &Development &Test \\
\midrule
\midrule
\makecell[l]{Isolation or\\loss of relationship}     & 2,486 & 923 & 913 \\\midrule
Transition of care    & 4,126 & 1,469 & 1,424 \\\midrule
Barriers to care      & 684  & 206 & 189 \\\midrule
Financial or job insecurity  & 2,119 & 652 & 725 \\\midrule
Housing instability   & 3,306 & 1,171 & 1,112 \\\midrule
Food insecurity       & 198 & 93 & 71 \\\midrule
Violence              & 1,277 & 483 &513 \\\midrule
Legal problems        & 1,494 & 509 & 566 \\\midrule
Substance abuse       & 3,984 & 1,251 & 1,420 \\\midrule
\makecell[l]{Psychiatric\\symptoms or disorders} & 4,630 & 1,649 & 1,472\\\midrule
Pain\tnote{*}                  & 1,128 & 337 & 317 \\\midrule
Patient disability    & 2,433 & 654 & 719 \\\midrule\midrule
Total & 27,865 & 9,397 &9,441  \\
\bottomrule
\end{tabular}
\begin{tablenotes}
    \item[*] Not an SBDH.
\end{tablenotes}
\end{threeparttable}
\caption{SBDH annotation distribution of VA-SBDH for the NER task.}
\label{tbl_va_sbdh_ctg}
\end{table}

\begin{table}
\fontsize{8pt}{8pt}\selectfont
\centering
\begin{threeparttable}
\begin{tabular}{lrrr}
\toprule
SBDH categories & Training &Development &Test \\
\midrule
\midrule
Social isolation \\
\hspace{4mm} Presence = yes      & 923 & 158 & 275  \\
\hspace{4mm} Presence = no      & 23 & 1 & 6  \\\midrule
Physical isolation    \\
\hspace{4mm} Presence = yes      & 318 & 46 & 84  \\
\hspace{4mm} Presence = no      & 2 & 0 & 0  \\\midrule
Transition of care   \\
\hspace{4mm} Presence = yes      & 613 & 82 & 152  \\
\hspace{4mm} Presence = no      & 5 & 0 & 1  \\\midrule
Barriers to care     \\
\hspace{4mm} Presence = yes      & 387 & 54 & 106  \\
\hspace{4mm} Presence = no      & 19 & 6 & 5  \\\midrule
Financial insecurity \\
\hspace{4mm} Presence = yes      & 801 & 100 & 233  \\
\hspace{4mm} Presence = no      & 16 & 1 & 2  \\\midrule
Job insecurity      \\
\hspace{4mm} Presence = yes      & 993 & 124 & 298  \\
\hspace{4mm} Presence = no      & 25 & 2 & 5  \\\midrule
Loss of relationship  \\
\hspace{4mm} Presence = yes      & 660 & 103 & 185  \\
\hspace{4mm} Presence = no      & 4 & 0 & 1  \\\midrule
Housing insecurity  \\
\hspace{4mm} Presence = yes      & 663 & 100 & 182  \\
\hspace{4mm} Presence = no      & 33 & 1 & 11  \\\midrule
Food insecurity    \\
\hspace{4mm} Presence = yes      & 504 & 72 & 136  \\
\hspace{4mm} Presence = no      & 15 & 7 & 1  \\\midrule
Violence           \\
\hspace{4mm} Presence = yes      & 494 & 76 & 148  \\
\hspace{4mm} Presence = no      & 8 & 1 & 3  \\\midrule
Legal problems  \\
\hspace{4mm} Presence = yes      & 345 & 55 & 107  \\
\hspace{4mm} Presence = no      & 8 & 1 & 2  \\\midrule
Substance abuse \\
\hspace{4mm} Presence = yes      & 717 & 109 & 226  \\
\hspace{4mm} Presence = no      & 125 & 30 & 36  \\\midrule
\makecell[l]{Psychiatric \\symptoms or disorders} \\
\hspace{4mm} Presence = yes      & 1,082 & 134 & 312  \\
\hspace{4mm} Presence = no      & 6 & 1 & 0  \\\midrule
Pain \tnote{*} \\
\hspace{4mm} Presence = yes      & 624 & 98 & 199  \\
\hspace{4mm} Presence = no      & 1 & 2 & 1  \\\midrule
Patient disability  \\
\hspace{4mm} Presence = yes      & 603 & 79 & 185  \\
\hspace{4mm} Presence = no      & 5 & 0 & 2  \\\midrule\midrule
Total & 10,022 & 1,443 &2,904\tnote{$\dagger$}\\
\bottomrule
\end{tabular}
  \begin{tablenotes}
    \item[*] Not an SBDH.
    \item[$\dagger$] We removed 2 annotations with made-up categories by GPT-4, outside the 15 target categories. So this is 2 less than the total annotation count mentioned in section \ref{err}.
  \end{tablenotes}
\end{threeparttable}
\caption{SBDH annotation distribution of Synth-SBDH for the NER task.}
\label{tbl_synth_sbdh_ctg}
\vspace{-4mm}
\end{table}

\begin{table}
\fontsize{8pt}{8pt}\selectfont
\centering
\begin{threeparttable}
\begin{tabular}{lrrr}
\toprule
SBDH categories & Training &Development &Test \\
\midrule
\midrule
Transition of care    & 566 & 73 & 138\\\midrule
Barriers to care      & 375  & 51 & 100  \\\midrule
Financial or Job insecurity  & 1,437 & 190 & 424\\\midrule
\makecell[l]{Isolation or\\loss of relationship}  & 1,430 & 228 & 406\\\midrule
Housing insecurity   & 602 & 88 & 167\\\midrule
Food insecurity       & 488 & 69 & 131 \\\midrule
Violence              & 467 & 73 &139 \\\midrule
Legal problems        & 315 & 47 & 102 \\\midrule
Substance abuse       & 624 & 90 & 195\\\midrule
\makecell[l]{Psychiatric symptoms\\or disorders} & 1,010 & 127 & 278\\\midrule
Pain \tnote{*}                 & 615 & 94 & 190 \\\midrule
Patient disability    & 567 & 76 & 169 \\\midrule\midrule
Total & 8,496 & 1,206 &2,439\\
\bottomrule
\end{tabular}
  \begin{tablenotes}
    \item[*] Not an SBDH.
  \end{tablenotes}
\end{threeparttable}
\caption{SBDH category distribution of Synth-SBDH for the MLC task.}
\label{tbl_synth_sbdh_ctg_mlc}
\vspace{-4mm}
\end{table}

\section{Task and experiment details}
\begin{figure*}[hbt]
    \centering
    \includegraphics[width=.9\linewidth,trim={4cm 0cm 4cm 0cm},clip]{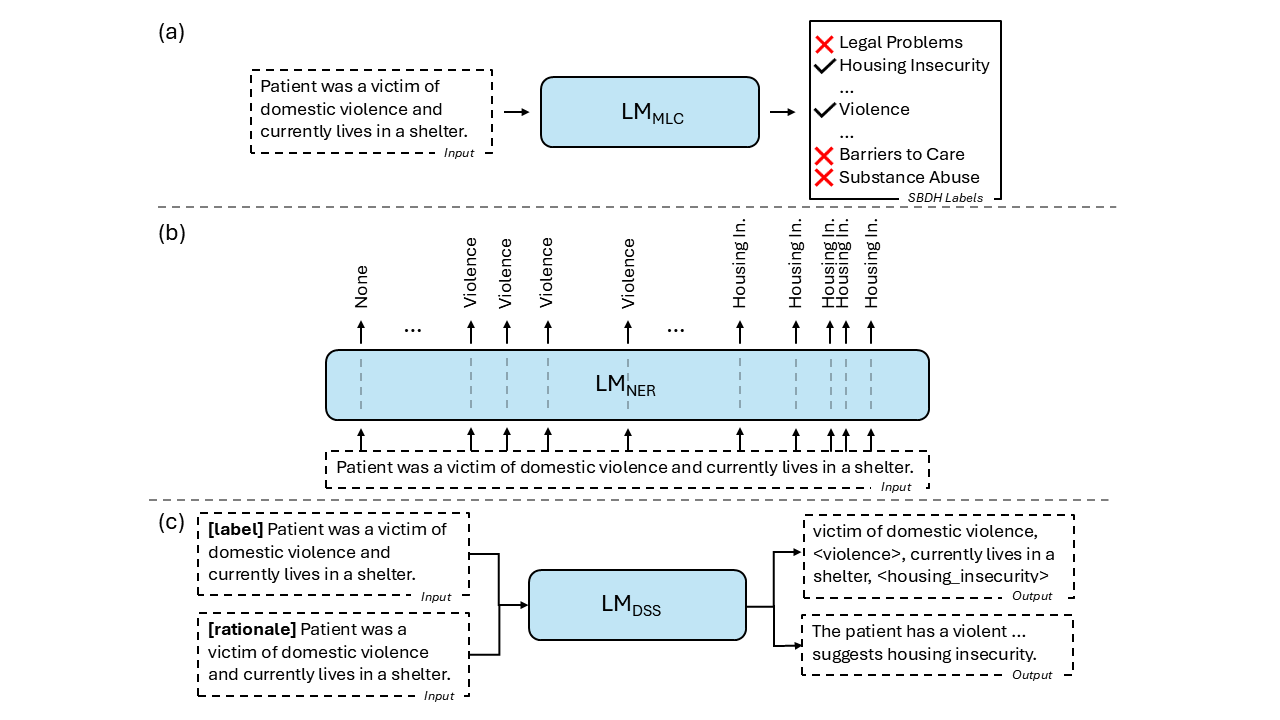}
    \caption{An illustration of the three tasks considered - (a) MLC, (b) NER, and (c) DSS.}
    \label{fig:tasks}
    \vspace{-2mm}
\end{figure*}
\subsection{Task details}
\label{app_task}
We considered three tasks as mentioned in section \ref{eval_tasks} (Figure \ref{fig:tasks}). For the MLC and NER tasks, we follow a three-step process. In the first step, we modify Synth-SBDH to align with the downstream task and dataset. We start by combining related SBDH categories in Synth-SBDH (as described in Appendix \ref{app_va_sbdh}) to have a total of 12  categories (instead of the original 15). For MLC, we convert Synth-SBDH and VA-SBDH into an MLC dataset by following these steps:
\begin{enumerate}
    \item Extract SBDH categories: We identify and extract all unique SBDH categories from the dataset.
    \item Aggregate: For each example in the dataset, we aggregate all the SBDH categories recognized within that example.
    \item Create Labels: We generate a set of labels for each example, where each label corresponds to an SBDH category present (\textit{presence}=`yes') in the example. An example can have multiple labels if it contains annotations of multiple SBDH categories.
    \item Binary Encoding: We encode the presence of each SBDH category in an example using a binary format (1 if present, 0 if not). This results in a multi-label format where each document is associated with a binary vector indicating the presence or absence of each named entity type.
\end{enumerate}
For NER, we convert examples from Synth-SBDH into BIO format for RoBERTa and clinicalRoBERTa models. For Seq2seq models, we rephrase NER as a generative task where given an example the models are trained to generate all text spans indicating any SBDH, each followed by the associated SBDH category. For DSS experiments, we reformulate SBDH extraction as the generative NER task described above. With multiple SBDH annotations, we concatenate them to create the rationale label. Ultimately, generative systems are trained in a multi-task learning framework to generate either the SBDH categories (along with related text spans) or rationales depending on the input. More details about DSS framework are available in the original paper \citep{hsieh2023distilling}. For all three tasks, we avoid nested annotations\footnote{For nested annotations, i.e. multiple annotations with overlapping text spans, we choose the first annotation.} and only consider annotations with \textit{presence}=`yes'.

In the second and third steps, we do a two-stage supervised fine-tuning (SFT). In the second step, we fine-tune models on the modified Synth-SBDH dataset (SFT$_{\text{stage1}}$), and in the third and last step, we use trained models from SFT$_{\text{stage1}}$ to further fine-tune on the task-specific real-world datasets (SFT$_{\text{stage2}}$). Note that for DSS, there is no other real-world SBDH dataset with rationales, so there is no SFT$_{\text{stage2}}$. Instead, we report the results on the expert-reviewed Synth-SBDH test set.

We adopted this two-stage SFT to better reflect real-world deployment scenarios. Specifically, we envision models trained on Synth-SBDH at SFT$_{\text{stage1}}$ being made publicly available so that healthcare personnel, particularly at resource-limited or remote facilities, can fine-tune them on relatively small amounts of in-domain, human-annotated data ((SFT$_{\text{stage2}}$)). This approach lowers the barrier for developing and deploying customized SBDH systems with limited data and computational resources. Moreover, the rationale aligns with widely accepted practices in SFT with synthetic data: initially training on synthetic data helps the model adapt to the target task, especially when labeled real-world data is scarce. Subsequent fine-tuning on real data further improves performance by grounding the model in domain-specific, high-quality annotations. This sequential strategy helps mitigate potential biases or errors introduced during the synthetic pretraining phase.

\subsection{Training configurations}
\label{app_hp}
For all experiments, we use the development sets of respective datasets to choose the best hyperparameter configurations. More details for training on downstream tasks are available in Table \ref{tbl_hp}. For NER and MLC, we only show configurations for SFT$_{\text{stage2}}$.
For MLC in SFT$_{\text{stage1}}$, we use the same configurations with the only change in num epoch, which is set to 40 (20 for Llama 3 8B). For NER in SFT$_{\text{stage1}}$, there are a few changes: batch size = 32, max seq length = 512, and GPU specs = 1 Tesla V100 32 GB.

\begin{table}[ht]
\fontsize{6.5pt}{6.5pt}\selectfont
\centering
\begin{threeparttable}
\begin{tabular}{lrrr}\toprule
 &MLC &NER &DSS \\\midrule
 Hyperparameters\\
\hspace{2mm} optimizer &AdamW &AdamW &AdamW \\
\hspace{2mm} beta1 &0.9 &0.9 &0.9 \\
\hspace{2mm} beta2 &0.95 &0.95 &0.999\\
\hspace{2mm} batch size &32 &16/8 &32 \\
\hspace{2mm} num epoch &8\tnote{a} &8 &100\tnote{b}\\
\hspace{2mm} weight decay &0.01 &0 &0\\
\hspace{2mm} warmup ratio &0.1/0.15 &0/0.1 &0\\
\hspace{2mm} gradient acc. steps &8 &1 &2 \\
\hspace{2mm} learning rate (lr) &1e-5/3e-4 &1e-5/5e-5 &5e-5 \\
\hspace{2mm} lr scheduler &linear/constant &linear &linear\\
\hspace{2mm} max seq length &256 &512 &256\\
GPU\\
\hspace{2mm} Count & 1 & 1 & 1\\
\hspace{2mm} Model & Tesla V100\tnote{c} & Tesla P40 & Tesla V100\tnote{c}\\
\hspace{2mm} Memory & 32GB & 24GB & 32GB\\
Compute time (approx.) &35mins/25mins\tnote{d} &3.5hrs/12-17hrs &2-3hrs\tnote{d}\\
\bottomrule
\end{tabular}
\begin{tablenotes}
    \item[a] 4 for Llama models.
    \item[b] 25 for FLAN-T5-xl.
    \item[c] Nvidia A100 80GB for LLMs.
    \item [d] 2-8hrs for LLMs.
\end{tablenotes}
\end{threeparttable}
\caption{Training configurations on downstream datasets. `/' separates values for RoBERTa-variants and Seq2seq models. For the DSS experiment, we have only Seq2seq models.}
\label{tbl_hp}
\end{table}

\section{More experiments}
\label{mlc_more_exp}

\subsection{MLC on VA-SBDH}
\label{mlc_va}
We report the results of MLC on VA-SBDH in Table \ref{result_ml_synth_vs_mimic}. We see 11.51\% improvements in macro F (72.38\% to 80.71\%) for RoBERTa with standard fine-tuning and 2.22\% with prompt-based fine-tuning. For  ClinicalRoBERTa the macro F improvements are 12.21\% (with standard fine-tuning) and 1.67\% (with prompt-based fine-tuning). Similar to MIMIC-SBDH$_\text{aligned}$, we also note substantially higher macro F score improvements than micro F scores. However, there was no noticeable gain for mamba models.

\begin{table}[hbt]
  \fontsize{7pt}{7.5pt}\selectfont
  \centering
  \begin{threeparttable}
  \begin{tabular}{lcc}
   \toprule
    Model &Micro F &Macro F \\\midrule
    Standard Fine-tuning\\\midrule
    RoBERTa-base & 81.16 ± 0.56	 & 72.38 ± 0.60 \\
    \hspace{2mm} With Synth-SBDH & \textbf{84.06 ± 0.08}  & \textbf{80.71 ± 0.15} \\
    \cdashline{1-3}
    ClinicalRoBERTa-base & 80.79 ± 0.17 & 71.69 ± 1.45  \\
    \hspace{2mm} With Synth-SBDH & \textbf{83.72 ± 0.03}  & \textbf{80.44 ± 0.15} \\
    \midrule
    Prompt-based Fine-tuning\\ \midrule
     RoBERTa-base & 84.53 ± 0.07	& 80.80 ± 0.21  \\
    \hspace{2mm} With Synth-SBDH & \textbf{85.39 ± 0.30}   & \textbf{82.59 ± 0.33}\\
    \cdashline{1-3}
    ClinicalRoBERTa-base & 84.12 ± 0.18	& 80.14 ± 0.43	    \\
    \hspace{2mm} With Synth-SBDH & \textbf{84.94 ± 0.29}	 & \textbf{81.48 ± 0.53}\\
    \cdashline{1-3}
    Mamba-130m & \textbf{85.11 ± 0.31}	& \textbf{81.96 ± 0.11} \\
    \hspace{2mm} With Synth-SBDH & 84.96 ± 0.09   & 81.66 ± 0.17 \\
    \cdashline{1-3}
    ClinicalMamba-130m  & 85.40 ± 0.06	& 82.41 ± 0.15 \\
    \hspace{2mm} With Synth-SBDH & \textbf{85.55 ± 0.24}   & \textbf{82.44 ± 0.35}\\
    \bottomrule
  \end{tabular}
  \end{threeparttable}
  \caption{SBDH detection as an MLC task on VA-SBDH when fine-tuned on MIMIC-SBDH$_\text{aligned}$ vs fine-tuned on Synth-SBDH. Fine-tuning baseline models on Synth-SBDH before fine-tuning on the target dataset generally yields performance improvements. Replacing Synth-SBDH with MIMIC-SBDH$_\text{aligned}$ degrades performance in all cases. Each cell value indicates the mean and standard deviation over three independent runs.}
  \label{result_ml_synth_vs_mimic}
  \vspace{-4mm}
\end{table} 

\subsection{Synth-SBDH vs MIMIC-SBDH$_\text{aligned}$}
\label{mlc_synth_vs_mimic}
In this experiment, we compare the utility of Synth-SBDH in comparison to the real-world dataset MIMIC-SBDH$_\text{aligned}$ by using them interchangeably in the training process. Specifically, during the SFT$_{\text{stage1}}$, we substitute Synth-SBDH with MIMIC-SBDH$_\text{aligned}$ and fine-tune all baseline models for an equivalent number of training steps (50 epochs for 4,917 training examples versus 40 epochs for 6,136 examples of Synth-SBDH). The results are presented in Table \ref{result_ml_synth_vs_mimic}. Despite being synthetic, we can see that Synth-SBDH outperforms MIMIC-SBDH$_\text{aligned}$ in enhancing model performance across all baselines, particularly for standard fine-tuning. We hypothesize that this improvement is attributed to the well-balanced examples and more comprehensive SBDH coverage provided by Synth-SBDH.

\subsection{LLMs with Synth-SBDH}
\label{llm}
\subsubsection{MLC}
\label{mlc_llm}
We fine-tune two LLMs - Llama 3.2 3B \citep{llama3_2modelcard} and Llama 3 8B \citep{llama3modelcard} for the MLC task on both MIMIC-SBDH$_\text{aligned}$ and VA-SBDH using QLoRA \citep{dettmers2023qlora}. The results are reported in Table \ref{result_ml_llama}. We notice that Llama models achieve higher F-scores compared to all other models (Tables \ref{result_ml}, \ref{result_ml_synth_vs_mimic} and \ref{result_ml_llama}) in the no-Synth-SBDH setting. This outcome is expected, given that Llama 3 8B has 61.8-64.2$\times$ more parameters (24.7-25.6$\times$ for Llama 3.2 3B) than the other models and was pre-trained on a significantly larger and more comprehensive corpus. 

However, the introduction of Synth-SBDH leads to a more compelling outcome: smaller language models (trained via prompt-based fine-tuning) achieve performance comparable to LLMs. For example, on MIMIC-SBDH, RoBERTa achieved a macro-F score of 91.11\%, slightly outperforming Llama 3 8B (91.02\% macro-F), despite the latter being approximately 64 times larger. On VA-SBDH, these models achieved macro-F scores of 82.59\% and 86.89\%, respectively — only a 5.21\% increase despite Llama 3 8B requiring approximately 15 times more VRAM and incurring nearly twice the cost (based on current Amazon SageMaker pricing), even after applying 4-bit quantization. In real-world clinical settings where computational resources are limited, such cost-performance trade-offs often lead clinicians and administrators to favor smaller models over more resource-intensive LLMs.

With Synth-SBDH, LLMs exhibit no significant improvements in performance. We hypothesize that this limited impact is due to VA-SBDH being a bigger and more balanced real-world data than MIMIC-SBDH$_\text{aligned}$. Furthermore, the strong baseline performance of these LLMs may reflect their advanced understanding of SBDH-related concepts, possibly due to the presence of clinical content in their pretraining corpus. Nonetheless, prior studies have shown that with sufficient synthetic data, it is possible to further enhance the performance of LLMs on downstream tasks \citep{wang2024notechat,tran2024bioinstruct}. Therefore, we believe that scaling up the Synth-SBDH dataset could potentially yield substantial gains for LLMs.

\begin{table}
  \fontsize{8pt}{10pt}\selectfont
  \centering
  \begin{threeparttable}
  \begin{tabular}{lcc}
    \toprule
    \multirow{2}{*}{Dataset} &\multicolumn{2}{c}{Prompt-based Fine-tuning} \\\cmidrule{2-3}
     &Micro F &Macro F \\\midrule
    MIMIC-SBDH$_\text{aligned}$\\\midrule
    Llama 3.2 3B\\
    \hspace{2mm}W/o Synth-SBDH   & \textbf{90.09 ± 0.37}	& \textbf{88.59 ± 1.24}  \\
    \hspace{2mm}With Synth-SBDH    & 89.78 ± 0.20   & 88.33 ± 0.74\\
    \cdashline{1-3}
    Llama 3 8B\\
    \hspace{2mm}W/o Synth-SBDH   & 91.99 ± 0.23	& 90.94 ± 0.43  \\
    \hspace{2mm}With Synth-SBDH    & \textbf{92.27 ± 0.25}   & \textbf{91.02 ± 0.65}\\
    \cdashline{1-3}
    RoBERTa-base\\
    \hspace{2mm}W/o Synth-SBDH   & 79.32 ± 0.92	& 55.64 ± 1.30  \\
    \hspace{2mm}With Synth-SBDH    & \textbf{91.11 ± 0.07}   & \textbf{91.11 ± 0.15}\\
    \midrule\midrule
    VA-SBDH\\\midrule
    Llama 3.2 3B\\
    \hspace{2mm}W/o Synth-SBDH    & 88.08 ± 0.16	& \textbf{86.20 ± 0.21}	    \\
    \hspace{2mm}With Synth-SBDH    & \textbf{88.19 ± 0.10}	 & 86.19 ± 0.21\\
    Llama 3 8B\\
    \hspace{2mm}W/o Synth-SBDH    & \textbf{88.71 ± 0.11}	& \textbf{87.07 ± 0.10}	    \\
    \hspace{2mm}With Synth-SBDH    & 88.49 ± 0.07	 & 86.89 ± 0.08\\
    \cdashline{1-3}
    RoBERTa-base\\
    \hspace{2mm}W/o Synth-SBDH   & 84.53 ± 0.07	& 80.80 ± 0.21  \\
    \hspace{2mm}With Synth-SBDH    & \textbf{85.39 ± 0.30}   & \textbf{82.59 ± 0.33}\\
    \bottomrule
  \end{tabular}
  \end{threeparttable}
  \caption{SBDH detection as an MLC task on MIMIC-SBDH$_\text{aligned}$ and VA-SBDH for LLMs. Each cell value indicates the mean and standard deviation over three independent runs.}
  \label{result_ml_llama}
\end{table}

\subsubsection{DSS}
\label{dss_llm}
Prior studies have shown that decoder-only models (e.g. Llama-series models) are not well-suited for sequence labeling tasks such as NER \citep{zaratiana2023gliner,wang2023instructuie}. Since we reframe SBDH extraction as an NER task in the DSS setting, we fine-tune FLAN-T5-xl \citep{chung2022scaling} , a 2.85B-parameter Seq2seq  model. The results are shown in Table \ref{result_dss_ext}. Similar to SLMs, We see that FLAN-T5-xl also achieves a significant performance boost with rationales - 1.23\% and 1.70\% increases in micro and macro F scores respectively. Interestingly, increasing model size (FLAN-T5-base to FLAN-T5-xl) did not yield any performance boost.

\begin{table}
  \fontsize{8pt}{10pt}\selectfont
  \centering
  \begin{threeparttable}
  \begin{tabular}{lccc}
    \toprule
    Model &Micro F &Macro F \\\midrule
    T5-base \\
    \hspace{2mm} Standard Fine-Tuning & 57.65 ± 0.30    & 54.94 ± 0.02   \\
    \hspace{2mm} DSS & \textbf{58.36 ± 0.26}    & \textbf{55.60 ± 0.26}  \\
    \hspace{2mm} Gain, $\Delta$ & 1.23\%    & 1.20\%  \\
    \cdashline{1-4}
    FLAN-T5-base \\ 
    \hspace{2mm} Standard Fine-Tuning & 57.19 ± 0.44    & 54.36 ± 0.50 \\
    \hspace{2mm} DSS & \textbf{57.70 ± 0.89}    & \textbf{54.94 ± 0.84} \\
    \hspace{2mm} Gain, $\Delta$ & 0.89\%    & 1.07\%  \\
    \cdashline{1-4}
     FLAN-T5-xl \\ 
    \hspace{2mm} Standard Fine-Tuning & 56.98±0.35	& 53.67±0.65 \\
    \hspace{2mm} DSS & \textbf{57.68±0.18}    & \textbf{54.58±0.57} \\
    \hspace{2mm} Gain, $\Delta$ & 1.23\%    & 1.70\%  \\
    \bottomrule
  \end{tabular}
  \end{threeparttable}
  \caption{SBDH extraction with DSS on Synth-SBDH. This is an extension of table \ref{result_dss} with LLM (FLAN-T5-xl). All models with DSS outperformed standard fine-tuning. Models are evaluated on the expert-reviewed test set. Each cell value indicates the mean and standard deviation over three independent runs.}
  \label{result_dss_ext}
  \vspace{-5mm}
\end{table}

\begin{table*}
  \fontsize{8pt}{10pt}\selectfont
  \centering
  \begin{threeparttable}
  \begin{tabular}{lcccc}
    \toprule
    \multirow{2}{*}{Model} &\multicolumn{2}{c}{Exact Matching} &\multicolumn{2}{c}{Relaxed Matching} \\\cmidrule{2-5}
    &Micro F &Macro F &Micro F &Macro F \\\midrule
    Encoder-only models\\\midrule
    RoBERTa-base & 71.09 ± 0.11    & \textbf{67.99 ± 0.19}   & 80.88 ± 0.19   & \textbf{78.22 ± 0.14} \\
    \hspace{2mm} + Fine-tuned on Synth-SBDH & \textbf{71.18 ± 0.12}    & 67.87 ± 0.25   & \textbf{81.00 ± 0.04}   & 78.08 ± 0.05\\
    \cdashline{1-5}
    ClinicalRoBERTa-base & 69.95 ± 0.12   & 66.03 ± 0.27   & 79.82 ± 0.16   & 76.36 ± 0.18 \\
    \hspace{2mm} + Fine-tuned on Synth-SBDH & \textbf{70.33 ± 0.11}    & \textbf{66.33 ± 0.39}   & \textbf{80.22 ± 0.07} & \textbf{76.61 ± 0.19}\\
    \midrule
    Seq2Seq models\\ \midrule
    T5-base & 28.28 ± 19.93    & 25.44 ± 18.24   & 31.81 ± 22.83   & 29.12 ± 21.20 \\
    \hspace{2mm} + Fine-tuned on Synth-SBDH & \textbf{64.70 ± 0.77}    & \textbf{60.73 ± 0.94}   & \textbf{74.59 ± 0.35}   & \textbf{71.13 ± 0.61} \\
    \cdashline{1-5}
    FLAN-T5-base & 68.77 ± 0.11    & 64.93 ± 0.13   & 78.44 ± 0.14   & 74.59 ± 0.26 \\
    \hspace{2mm} + Fine-tuned on Synth-SBDH & \textbf{69.59 ± 0.23}    & \textbf{66.34 ± 0.31}   & \textbf{78.92 ± 0.49}   & \textbf{75.62 ± 0.51}\\
    \bottomrule
  \end{tabular}
  \end{threeparttable}
  \caption{SBDH detection as an NER task on VA-SBDH with relaxed matching. This is an extension of table \ref{result_ner}. Fine-tuning baseline models on Synth-SBDH before fine-tuning on the target dataset improves both exact and relaxed matching performance in almost all cases. Each cell value indicates the mean and standard deviation over three independent runs.}
  \label{result_ner_relaxed}
\end{table*}

\section{Human and LLM evaluation}
\label{eval}


\begin{table*}
\fontsize{8pt}{8pt}\selectfont
\centering
\begin{threeparttable}
\begin{tabular}{lllllll}
\toprule
SBDH categories & \#Annotations   & \textit{Keep} &\textit{Update} &\textit{Discard} &\textit{Add} \\
\midrule
\midrule
Social isolation      & 403 & 274 (67.99\%) & 35 (8.69\%) &  0 (0\%) & 94 (23.33\%) \\\midrule
Physical isolation    & 84 & 38 (45.24\%) & 18 (21.43\%) & 10 (11.90\%) & 18 (21.43\%) \\\midrule
Transition of care    & 169 & 114 (67.46\%) & 9 (5.33\%) & 25 (14.79\%) & 21 (12.43\%) \\\midrule
Barriers to care      & 136  & 89 (65.44\%) & 3 (2.21\%) & 18 (13.24\%) & 26 (19.12\%)   \\\midrule
Financial insecurity  & 270 & 212 (78.52\%) & 10 (3.70\%) & 3 (1.11\%) & 45 (16.67\%)\\\midrule
Job insecurity        & 347 & 287 (82.71\%) & 13 (3.75\%) & 1 (0.29\%) & 46 (13.26\%)\\\midrule
Loss of relationship  & 258 & 133 (51.55\%) & 51 (19.77\%) & 3 (11.63\%) & 71 (27.52\%)\\\midrule
Housing insecurity   & 275 & 178 (64.73\%) & 29 (10.55\%) & 0 (0\%) & 68 (24.73\%)\\\midrule
Food insecurity       & 153 & 123 (80.39\%) & 3 (1.96\%) & 12 (7.84\%) & 15 (9.80\%)\\\midrule
Violence              & 201 & 128 (63.68\%) &25 (12.44\%) & 4 (1.99\%) & 44 (21.89\%)\\\midrule
Legal problems        & 128 & 95 (74.22\%) & 15 (11.72\%) & 1 (0.78\%) & 17 (13.28\%)\\\midrule
Substance abuse       & 277 & 210 (75.81\%) & 38 (13.72\%) & 7 (2.53\%) & 22 (7.94\%) \\\midrule
Psychiatric   symptoms or disorders & 367 & 267 (72.75\%) & 13 (3.54\%) & 29 (7.90\%)& 58 (15.80\%)\\\midrule
Pain \tnote{*}                 & 230 & 196 (85.22\%) & 13 (5.65\%) & 1 (0.43\%) & 20 (8.70\%) \\\midrule
Patient disability    & 187 & 134 (71.66\%) & 11 (5.88\%) & 26 (13.90\%) & 16 (8.56\%) \\\midrule
Others\tnote{$\dagger$}       & 2 & 0 & 0 & 2 & 0\\\midrule\midrule
Total & 3,487 & 2,478 (71.06\%) &286 (8.20\%) & 142 (4.07\%) & 581 (16.66\%) \\
\bottomrule
\end{tabular}
\begin{tablenotes}
    \item[*] Not an SBDH.
    \item[$\dagger$] Instances where GPT-4 created new SBDH categories.
\end{tablenotes}
\end{threeparttable}
\caption{Category-wise breakdown of human evaluation for all 14 SBDH Categories. The percentages inside parentheses in each row sum up to ~100\% (there might be rounding errors).}
\label{sbdh_ctg_eval}
\end{table*}

\subsection{Annotator characteristics}
Our annotators are native English speakers with extensive experience in electronic health record (EHR) annotation. Each has a minimum of eight years of experience in EHR annotation and at least four years specifically in annotating SBDH within clinical notes. Both hold bachelor's degrees in Biology, with one also possessing a Master’s degree in Public Health. They both reside in the United States and are co-authors of this study. Compensation for their annotation and evaluation work was provided at a rate of \$40 per hour.

\subsection{Annotation review guidelines}
\begin{enumerate}
    \item Follow the SBDH definitions as described in Table \ref{synth_sbdh_ctg} and examples as provided in Appendix \ref{app_seed}. Here are some category-specific instructions-
    \begin{itemize}
        \item  Financial Insecurity: Consider references to adequate health insurance, and the presence of a pension as evidence of financial \textbf{security}.
        \item Psychiatric Symptoms or Disorders: References to Alzheimer’s disease/other dementia, memory issues, stress, sadness, and worry \textbf{not} included in this category. An exception to this is the mention of more generic stress where the cause can not be determined from the context. Such mentions of stress should be considered under this category.
    \end{itemize}
    \item If a sentence contains text spans belonging to multiple SBDH categories, each text span should be annotated and assigned to its corresponding category. Take the following example-
    \begin{tcolorbox}
        \textbf{Text}: The veteran suffers from PTSD due to combat experiences and is socially isolated.\\
        \textbf{Textspan}: suffers from PTSD || combat experiences || socially isolated\\
        \textbf{SBDH}: Psychiatric Symptoms or Disorders || Violence || Social Isolation\\
        \textbf{Presence}: yes || yes || yes\\
        \textbf{Period}: current || history || current
    \end{tcolorbox}
    \item If a text span contains elements of more than one SBDH category, and GPT-4 chose one of those categories, consider that as a correct annotation. However, in case of adding a new annotation,   prioritize the more specific category. Here is an example-
    \begin{tcolorbox}
        \textbf{Text}: Patient has been struggling to pay mortgage and is at risk of foreclosure.\\
        \textbf{Textspan}: struggling to pay mortgage || at risk of foreclosure\\
        \textbf{SBDH}: Financial Insecurity || Housing Insecurity\\
        \textbf{Presence}: yes || yes \\
        \textbf{Period}: current || current
    \end{tcolorbox}
    The text span `struggling to pay mortgage’ indicates both financial and housing insecurity; so this is a correct annotation by GPT-4. However, if this annotation is missing and needs to be added, categorize it as housing insecurity.
    \item When possible, avoid nested annotations. However, for situations where nested annotations are necessary, consider them as such. The following example demonstrates such a case.
    \begin{tcolorbox}
        \textbf{Text}: Patient is unable to pay for her medications and is worried about her financial situation.\\
        \textbf{Textspan}: unable to pay for her medications || worried about her financial situation || unable to pay for her medications\\
        \textbf{SBDH}: Barriers to Care || Financial Insecurity || Financial Insecurity\\
        \textbf{Presence}: yes || yes || yes\\
        \textbf{Period}: current || current || current
    \end{tcolorbox}
    Here `unable to pay for her medications' indicates both `Barriers to Care' and `Financial Insecurity' categories and  both should be annotated.
    \item If there is not enough context to detect ‘period’, use ‘unclear’.
    \item Consider each annotation for one of these four actions - 
    \begin{enumerate}
        \item \textit{Keep}: Keep the annotation if it is correct. In the following example, the text span `broke up with her boyfriend’ was categorized as `loss of relationship', and the text span `signs of depression’ was categorized as `psychiatric symptoms or disorders' by GPT-4, both of which are correct according to the SBDH definitions from Table \ref{synth_sbdh_ctg}.
        \begin{tcolorbox}
            \textbf{Text}: Patient broke up with her boyfriend recently and she shows signs of depression. \\
            \textbf{Textspan}: broke up with her boyfriend || signs of depression \\
            \textbf{SBDH}: Loss of Relationship || Psychiatric Symptoms or Disorders 
        \end{tcolorbox}
        \item \textit{Update}: Update to correct the annotation if there is an error in judgment. In the following example, GPT-4 categorized ‘drug trafficking’ as substance abuse. Drug trafficking belongs to the legal problems category.
        \begin{tcolorbox}
            \textbf{Text}: Patient was in prison for five years due to drug trafficking. \\
            \textbf{Textspan}: drug trafficking \\
            \textbf{SBDH}: Substance Abuse 
        \end{tcolorbox}
        \item \textit{Discard}: Discard the annotation if the annotation is unnecessary or erroneous. In the following example, GPT-4 labeled `involved in a car accident’ as patient disability with rationale - `Patient was involved in an accident, which might have caused physical trauma or injuries'. However, this is a far-fetched inference and not supported by the content.
        \begin{tcolorbox}
            \textbf{Text}: Patient has been involved in a car accident recently and is experiencing injuries. \\
            \textbf{Textspan}: involved in a car accident \\
            \textbf{SBDH}: Patient Disability 
        \end{tcolorbox}
        \item \textit{Add}: Add SBDH annotations missed by GPT-4. In the following example, GPT-4 annotated the spans `living in a homeless shelter’ (housing insecurity) and `loss of her job’ (job insecurity) but did not include `could not afford rent’ (housing insecurity).
        \begin{tcolorbox}
            \textbf{Text}: Patient has been living in a homeless shelter as she could not afford rent due to loss of her job. \\
            \textbf{Textspan}: living in a homeless shelter || loss of her job\\
            \textbf{SBDH}: Housing insecurity || Job Insecurity
        \end{tcolorbox}
    \end{enumerate}
\end{enumerate}

\subsection{Rationale rating}
We used a 4-point Likert scale to rate all GPT-4 generated rationales in the test set to assess their quality. The scale is as follows -  
\begin{itemize}
    \item \textit{Incorrect} (1 pt.): The rationale provided is entirely incorrect and does not align with the context or definitions provided in the prompt.
    \item \textit{Incorrect with Direct Inference} (2 pt.): The rationale contains some elements of inference related to an SBDH, making it a convincing explanation or referring to another annotation, but overall, is incorrect.
    \item \textit{Correct with Unnecessary or Irrelevant or Incomplete Information} (3 pt): The rationale is correct in detecting the SBDH and other attributes (presence and/or period) but includes additional information that - 
    \begin{enumerate}
        \item is not supported by the context, or
        \item is logically incorrect, or
        \item does not contribute to the SBDH detection thought process, or
        \item misses important information such as mention of the concept or topic.
    \end{enumerate}
    \item \textit{Correct} (4 pt.): The rationale provided is entirely correct without any unnecessary or irrelevant information.
\end{itemize}
For rationales with a rating of less than 4, experts provide a \textit{Correct} rationale. The silver-level test set of Synth-SBDH contains only \textit{Correct} rationales.

\subsection{Interesting observations}
\label{eval_obs}
Here we list a few interesting cases found during the human evaluation.
\begin{itemize}
    \item GPT-4 assumed any instances of suicidal behavior should be categorized under the `Psychiatric Symptoms or Disorders' category. Considering no instruction pertaining to the categorization of suicidal behavior was provided to GPT-4, this SBDH category is the best fit.
    \item GPT-4 linked social worker’s assistance to `Barriers to Care' category despite not having such an example in the prompt. GPT-4 did a good job at extrapolating this into the `Barriers to Care' category. Here is an example.
    \begin{tcolorbox}
        \textbf{Text}: The patient was very happy with the social worker's assistance in understanding her prescription plan.\\
        \textbf{Textspan}: \textcolor{blue}{social worker's assistance}\\
        \textbf{SBDH}: \textcolor{blue}{Barriers to Care} \\
        \textbf{Presence}: \textcolor{blue}{no} \\
        \textbf{Period}: \textcolor{blue}{current} \\
        \textbf{Reasoning}: \textcolor{blue}{The patient was able to understand their prescription plan with the help of a social worker which suggests absence of barriers to care.}
    \end{tcolorbox}
    \item GPT-4 extrapolated that the use of a guide dog indicates a disability which is an interesting assumption.
    \begin{tcolorbox}
        \textbf{Text}: The patient is blind and relies on a guide dog for mobility.\\
        \textbf{Textspan}: blind || \textcolor{blue}{relies on a guide dog for mobility}\\
        \textbf{SBDH}: Patient Disability || \textcolor{blue}{Patient Disability} \\
        \textbf{Presence}: yes || \textcolor{blue}{yes} \\
        \textbf{Period}: current || \textcolor{blue}{current} \\
        \textbf{Reasoning}: Blindness is a disability. || \textcolor{blue}{Reliance on a guide dog is a sign of disability.}
    \end{tcolorbox}
    
    
    \item GPT-4 had some difficulty distinguishing between substance abuse and legitimate use of prescribed controlled substances. Here is such an example where GPT-4 categorized ‘prescribed opioids’ as `Substance Abuse' though it was prescribed and not an illicit use. Note that there is also a missing annotation here (for text span `pain management').
    \begin{tcolorbox}
        \textbf{Text}: Veteran continues to have back pain and has been prescribed opioids for pain management.\\
        \textbf{Textspan}: back pain || \textcolor{red}{prescribed opioids}\\
        \textbf{SBDH}: Pain || \textcolor{red}{Substance Abuse} \\
        \textbf{Presence}: yes || \textcolor{red}{yes} \\
        \textbf{Period}: current || \textcolor{red}{current} \\
        \textbf{Reasoning}: The patient is experiencing physical discomfort. || \textcolor{red}{The patient has been provided with prescription medication for pain management.}
    \end{tcolorbox}
    \item At times, GPT-4 reached an illogical conclusion not supported by the available context.  In the following example, GPT-4 categorized the span ‘worsening symptoms' as `Psychiatric Symptoms or Disorders' despite the lack of context to indicate that symptoms are psychiatric in nature. 
    \begin{tcolorbox}
        \textbf{Text}: Patient was readmitted after discharge due to worsening symptoms.\\
        \textbf{Textspan}: readmitted after discharge || \textcolor{red}{worsening symptoms}\\
        \textbf{SBDH}: Transitions of Care || \textcolor{red}{Psychiatric Symptoms or Disorders} \\
        \textbf{Presence}: yes || \textcolor{red}{yes} \\
        \textbf{Period}: current || \textcolor{red}{current} \\
        \textbf{Reasoning}: Readmission soon after discharge indicates a transition of care. || \textcolor{red}{Worsening symptoms indicate underlying health issues.}
    \end{tcolorbox}
    \item GPT-4 rationales occasionally overreached or conflated concepts.
    \begin{tcolorbox}
        \textbf{Text}: Patient recently moved to a new city and is facing difficulties accessing regular healthcare.\\
        \textbf{Textspan}: \textcolor{red}{recently moved to a new city} || facing difficulties accessing regular healthcare\\
        \textbf{SBDH}: \textcolor{red}{Transitions of Care} || Barriers to Care \\
        \textbf{Presence}: \textcolor{red}{yes} || yes \\
        \textbf{Period}: \textcolor{red}{current} || current \\
        \textbf{Reasoning}: \textcolor{red}{Moving to a new city is a significant event that can potentially lead to changes in healthcare and therefore can be regarded as a transition in care.} || Difficulty in accessing healthcare is indicative of barriers to care.
    \end{tcolorbox}
    Here, GPT-4 categorized the text span ‘recently moved to a new city’ as `Transitions of care', with a vague and inappropriate rationale. Another interesting example is the following with the `Job insecurity' annotation.
    \begin{tcolorbox}
        \textbf{Text}: Patient recently divorced and now faces the challenge of caring for three children while working two jobs.\\
        \textbf{Textspan}: recently divorced || \textcolor{red}{working two jobs}\\
        \textbf{SBDH}: Loss of Relationship || \textcolor{red}{Job Insecurity} \\
        \textbf{Presence}: yes || \textcolor{red}{yes} \\
        \textbf{Period}: current || \textcolor{red}{current} \\
        \textbf{Reasoning}: A recent divorce indicates a loss of relationship. || \textcolor{red}{Working two jobs may suggest job insecurity due to insufficient income from a single job.}
    \end{tcolorbox}
    The correct SBDH category here is `Financial insecurity'.
    \item Experts also noticed annotations where it was difficult to decide on \textit{period} from the synthetic examples but GPT-4 made a decision regardless, partly because the data generation prompt instructed GPT-4 to consider only `current' and `history' as the \textit{period} values.

\end{itemize}
\subsection{Prompt for LLM evaluation}
To conduct LLM evaluation, we ask GPT-4 to identify whether each annotation from Synth-SBDH is correct or not. Labeling an annotation as correct is analogous to the \textit{keep} action by human experts while the opposite is similar to either \textit{update} or \textit{discard} actions. Note that we did not ask GPT-4 to identify missing annotations. We used the following prompt to evaluate Synth-SBDH by GPT-4.

\begin{Verbatim}[breaklines=true]
The social determinants of health (SDOH) are the non-medical factors that influence health outcomes. They are the conditions in which people are born, grow, work, live, and age, and the wider set of forces and systems shaping the conditions of daily life. SDOHs have a major impact on people’s health, well-being, and quality of life. SDOHs encompass factors such as socioeconomic status, access to healthy food, education, housing, and physical environment, to name a few. Together with behavioral factors such as substance abuse, we get Social and behavioral determinants of health (SBDH). Below are 15 SBDH categories with definitions that we will consider.

1. Food Insecurity: Lack of consistent access to enough food for every person in a household to live an active, healthy life.
2. Job Insecurity: Job insecurity includes unemployment, underemployment, unstable employment, fear of losing a job or benefits, and vocational rehabilitation/training.
3. Housing Insecurity: Housing insecurity refers to unstable housing due to a variety of reasons which may include eviction, inability to afford rent, foreclosure, or displacement due to domestic/roommate/landlord issues.
4. Financial Insecurity: The anxiety produced by the possible exposure to adverse economic events and the anticipation of the difficulty of recovering from them.
5. Legal Problems: Legal problems entail violations of law, associated punishments, and mention of related officials, places, and processes e.g., attorney, judge, parole officer, court, jail, prison, incarceration, child custody/child support issues.
6. Social Isolation: A state in which the individual lacks a sense of belonging socially, lacks engagement with others, has a minimal number of social contacts, and they are deficient in fulfilling and quality relationships.
7. Physical Isolation: Physical isolation results in less involvement with others, often due to disability, illness, housebound or bedbound, that prevents active participation in life outside of the home/immediate physical environment.
8. Loss of Relationship: A loss of a personal relationship, including divorce, separation, death, estrangement, or breakdown of interpersonal communication.
9. Barriers to Care: Barriers to care are factors that interfere with healthcare access, and may include transportation issues, cognitive or communication difficulties, lack of trust in the care system, or lack of rapport with provider(s).
10. Violence: The violence category includes elements of the individual\'s environment, as well as the larger societal environment. The presence of weapons, various types of abuse (physical, emotional/psychological, sexual), exposure to combat, bullying, harassment, threats, and racism are categorized as violence. Violence includes cases of both perpetrators and victims.
11. Transitions of Care: The transitions of care category identifies healthcare-related points of vulnerability; examples include admission, discharge, medication change, and change of provider.
12. Pain: The pain category considers acute and chronic pain, arthralgia, migraine, and evidence of pain through mention of pain management/mitigation.
13. Patient Disability: The patient disability category includes impairments that affect daily life as evidenced by the presence of assistive devices, disability payments, and military service-connected ratings.
14. Substance Abuse: Substance Abuse (marijuana excluded) covers the use of both legal (alcohol, tobacco) and illicit substances, addiction, substance abuse treatment/rehab/sobriety groups, and relapse.
15 Psychiatric Symptoms or Disorders: Psychiatric Symptoms or Disorders category includes emotional/psychological difficulties and conditions that affect the ability to function well in daily life.

Your task is to validate SBDH annotation for a text sequence that emulates the language a patient's electronic health records notes. The text sequence has six sections - `Text', `Textspan', `SBDH', `Presence', and `Period'. `Textspan' is a text span from the `Text' with indications of SBDH, `SBDH' indicates the annotated SBDH category, `Presence' indicates whether the SBDH is present or not - yes or no, and `Period' indicates the timeframe of the said SBDH - current (exists currently) or history (events from the past). Here are the steps for you to follow - 
1. Analyze the text sequence carefully and try to figure out all possible mentions of SBDH from different categories.
2. Analyze the provided annotation (`Textspan', `SBDH', `Presence', and `Period') and critique its validity.
3. Generate `yes' if the annotation is correct and `no' otherwise, put it under the `Correct' section.
4. Generate reasoning or rationale behind the annotation, and put it under the `Reasoning section. Mention what is wrong with the annotation if you find it incorrect.
5. Do not explicitly mention phrases such as `the annotation is correct/incorrect' in the `Reasoning' section, rather keep it concise and indicate what the `Textspan' indicates in the context of `Text'. 
6. Avoid single or double quotation marks.
7. Format your answer as a valid Python dictionary with the following structure: 
{
    `Correct':...,
    `Reasoning':...
}

Below are a few sample examples.

Example 1.
Text: Patient's Seroquel dose remains the same.
Textspan: dose remains the same
SBDH: Transitions of Care
Presence: no
Period: current
Reasoning: Seroquel dosage did not change, so no transition of care.

...

Example 5.
Text: Veteran is anticipated to discharge home once medically stable for discharge transported by his wife.
Textspan: discharge 
Reasoning: Discharge from the hospital is considered as transition of care. 
SBDH: Transitions of Care
Presence: yes 
Period: current 

Now evaluate the following example. 

{Single annotation from Synth-SBDH to evaluate}

\end{Verbatim}

\section{Intended use and potential misuse of Synth-SBDH}
\label{use}

The intended use of Synth-SBDH encompasses both research and deployable products. Researchers can utilize Synth-SBDH to develop and test NLP systems to extract SBDH in clinical text and conduct different clinical studies to analyze their associations or causality with different clinical or adversarial outcomes. For developers, the dataset offers opportunities to create or enhance applications, such as clinical decision support systems or health analytics tools, by integrating SBDH insights into operational environments.
However, users of the dataset should be mindful of potential misuse:
\begin{itemize}
    \item Overreliance on Synthetic Data: Synthetic data like Synth-SBDH may not capture the full complexity and variability of real-world data. Users should avoid using it as the sole basis for critical decisions or clinical practices. We recommend users to collect a reasonable sample of human-labeled data to facilitate domain adaptation for any system only fine-tuned on Synth-SBDH.
    \item Bias and Fairness: Synthetic data generated by LLMs may inherit biases from the training data and few-shot examples \citep{deletang2023language,navigli2023biases}. Users should be cautious of these biases and consider them when interpreting results or deploying models in diverse healthcare settings.
    \item Privacy Concerns: Though we tried to ensure that any example in Synth-SBDH does not resemble any real patient information, we urge users to ensure that no real patient data is being inadvertently exposed and adhere to ethical guidelines for data use.
    \item Ethical Use: Users should avoid using the dataset for purposes that could harm individuals or groups, such as creating discriminatory algorithms or spreading misinformation.
    \item Compliance with Legal and Regulatory Standards: Users must ensure that their use of synthetic data complies with relevant laws, regulations, and institutional policies, particularly in healthcare and data protection contexts.
\end{itemize}
We believe, by considering these factors, users can responsibly leverage the benefits of Synth-SBDH for advancing research and developing innovative applications while mitigating potential risks.

\end{document}